\documentclass{article}

\usepackage{arxiv}

\usepackage[utf8]{inputenc} 
\usepackage[T1]{fontenc}    
\usepackage[hidelinks]{hyperref}       
\usepackage{url}            
\usepackage{booktabs}       
\usepackage{amsfonts}       
\usepackage{nicefrac}       
\usepackage{microtype}      
\usepackage{lipsum}
\usepackage{graphicx}
\usepackage{academicons}
\usepackage[table]{xcolor}
\usepackage{siunitx}
\usepackage{algorithm}
\usepackage{algpseudocode}
\usepackage{tikz}
\usepackage{amsmath}
\usepackage{soul}
\usepackage{setspace}

\usepackage[
  backend=biber,
  doi=true,
  sorting=none,
  maxnames=99,
  minnames=1,
  maxcitenames=2,
  mincitenames=1,
  url=false]{biblatex}
\let\breakedtextcite\textcite
\renewcommand{\textcite}[1]{\mbox{\breakedtextcite{#1}}}

\newcommand{\orcid}[1]{}

\newcommand{\largearrow}[0]{
\begin{tikzpicture}
  \draw[white] (0, 0) -- (2cm, 1cm);
  \begin{scope}[scale=0.5, yshift=4cm]
    \fill[fill=white, draw=black] (0, 0.4) -- (3cm, 0.4cm) -- (3cm, 1cm) --
    (4cm, 0cm) -- (3cm, -1cm) -- (3cm, -0.4cm) -- (0cm, -0.4cm) -- cycle;
  \end{scope}
\end{tikzpicture}}

\newcommand{\smallarrow}[0]{
\begin{tikzpicture}
    \draw[white] (0, 0) -- (2cm, 1cm);
    \begin{scope}[scale=0.5, yshift=3cm]
      \fill[fill=white, draw=black] (0, 0.4) -- (3cm, 0.4cm) -- (3cm, 1cm) --
      (4cm, 0cm) -- (3cm, -1cm) -- (3cm, -0.4cm) -- (0cm, -0.4cm) -- cycle;
    \end{scope}
\end{tikzpicture}}

\graphicspath{{./figures}}

\ifdefined\doublestretch
\fi

\definecolor{good}{RGB}{255, 204, 20}
\definecolor{bad}{RGB}{200,50,90}

\title{
  Cross-sectional Topology Optimization of Slender Soft Pneumatic Actuators 
  using Genetic Algorithms and Geometrically Exact Beam Models}

\author{
 Leon Schindler\orcid{0009-0002-6530-5967} \\
  Institute for Computational Physics in Engineering \\
  RPTU Kaiserslautern-Landau\\
  Kaiserslautern, DE 67663 \\
  \texttt{leon.schindler@mv.rptu.de} \\
   \And
 Kristin Miriam de Payrebrune\orcid{0000-0003-4358-1960} \\
 Institute for Computational Physics in Engineering \\
  RPTU Kaiserslautern-Landau\\
  Kaiserslautern, DE 67663 \\
  \texttt{kristin.payrebrune@mv.rptu.de} \\
}

\begin{document}
\maketitle
\begin{abstract}
The design of soft robots is still commonly driven by manual trial-and-error approaches, requiring the manufacturing of multiple physical prototypes, which in the end, is time-consuming and requires significant expertise.
To reduce the number of manual interventions in this process, topology optimization can be used to assist the design process.
The design is then guided by simulations and numerous prototypes can be tested in simulation rather than being evaluated through laborious experiments.
To implement this simulation-driven design process, the possible design space of a slender soft pneumatic actuator is generalized to the design of the circular cross-section.
We perform a black-box topology optimization using genetic algorithms to obtain a cross-sectional design of a soft pneumatic actuator that is capable of reaching a target workspace defined by the end-effector positions at different pressure values.
This design method is evaluated for three different case studies and target workspaces, which were either randomly generated or specified by the operator of the design assistant.
The black-box topology optimization based on genetic algorithms proves to be capable of finding good designs under given plausible target workspaces.
We considered a simplified simulation model to verify the efficacy of the employed method.
An experimental validation has not yet been performed.
It can be concluded that the employed black-box topology optimization can assist in the design process for slender soft pneumatic actuators.
It supports at searching for possible design prototypes that reach points specified by corresponding actuation pressures.
This helps reduce the trial-and-error driven iterative manual design process and enables the operator to focus on prototypes that already offer a good viable solution.
\end{abstract}


\section{Introduction}
\label{sec:introduction}

Soft robotics is a novel subfield of continuum robots.
Two major features of soft robots compared to conventional rigid robots are the soft compliant material and their infinite degree-of-freedom actuation~\cite{Rus2015}.
These properties are, for example, beneficial in environments that are unstructured or where a collaboration with humans is intended~\cite{Rus2015}.
Although the continuous deformation is advantageous it is also challenging from a design and modelling perspective.
\textcite{Lipson2014} states that "[\dots] many engineers guide their design efforts using intuition fed from daily mechanical experiences regarding behavior of rigid kinematics and dynamics, such intuition is poor and qualitative at best when it comes to soft materials".
This effect is amplified by the lack of simulation tools, non-linear material laws and dynamic interactions~\cite{Lipson2014}.

The actuation principles of soft robots can be classified by the stimulation type; these include, among others, electrically, magnetically, thermally, or  pressure-stimulated actuators~\cite{El-Atab2020}.
In this study we focus on pressure-driven slender soft pneumatic actuators with a circular cross-section and arbitrary configurations of air chambers.
These air chambers can be assigned to one of three pressure supplies, which results in a spatial deformation of the actuator when pressurized.
While we neglect torsional effects, the spatial deformation can be described by elongation, bending and shearing.

Manual soft robotic design, influenced by bio-inspired organisms, might not yield optimal results given a specific task~\cite{Pinskier2022}.
However, there are methods that aim at improving these design mechanisms or to automate it entirely.
One such approach is presented in the present work, where we focus on the design optimization of slender soft pneumatic actuators using genetic algorithms.
Furthermore, we restrict the behavior of the slender soft pneumatic actuator to the deformation of its centerline.
This reduction decreases the computational complexity from a three-dimensional volumetric problem to a one-dimensional problem, which we solve using geometrically exact beam models.
A black-box topology optimization of the cross-sectional geometry of a slender soft-pneumatic actuator is performed to fit a target workspace spanned by the position of the end-effector at different actuation pressures.
Black-box topology optimizations can be used in cases where
no gradient information of the analyzed model is available~\cite{Guirguis2020}.
Although it is theoretically possible to compute gradient information for the mapping from cross-sectional geometry to the workspace defined by the end-effector positions under different input pressures, this would require a physical simulator that is differentiable with respect to the design parameters of the cross-section~\cite{Bacher2021} or numerical approximation of the gradients.
We perform the numerical simulation in a custom Python program, which, in its current implementation, is not automatically differentiable and, therefore, cannot be used in a differentiable simulation.

Due to the computational complexity of the non-linear problems that arise when optimizing soft robotic systems, the design optimization of such systems has only gained traction in the recent past~\cite{Pinskier2022}.
\textcite{Mosser2023} use B\'{e}zier curve-based genetic algorithms in combination with Gaussian mixture points to optimize the cavity structure of three-dimensional soft pneumatic actuators.
The simulation of the soft pneumatic actuator is performed using finite element models supported by a transfer-learned convolutional neural network (CNN).
\textcite{Zhang2017} performed a topology optimization using the solid isotropic material with penalization (SIMP) method to derive the three-dimensional shape of a soft gripper.
This paper is followed up by \textcite{Zhang2019c} where this approach is extended to a multi-material topology optimization of more generally applicable soft pneumatic actuators by combining hard and soft material.
The soft gripper in these articles is optimized to obtain a maximal tip deflection under a specified actuation pressure.
Furthermore, the design space is restricted to the outer shell of the soft pneumatic actuator, while the pressurized surface area stays constant.
\textcite{Caasenbrood2020} adapted a nonlinear topology optimization with pressure-dependent loads using the solid isotropic material with penalization method.
A different approach was chosen in \textcite{Chen2019a} where a bi-directional evolutionary optimization (BESO) is used to iteratively consider the pressure-dependent loads in the topology optimization problem.
Both articles focus on the maximum deflection at the end of a soft pneumatic network as an objective.
A design optimization of a cylindrical soft pneumatic actuator is presented in
\textcite{Runge2017b}.
This soft pneumatic actuator consists of three individual pressure chambers, which are modified via their geometrical properties (inner and outer radius, position and thickness) using a genetic algorithm to optimize the curvature of the actuated soft pneumatic.

A more generalized design of soft robots is considered with a voxel-based description.
\textcite{Hiller2010a, Hiller2010} use genetic algorithms with different level-set encodings to design multi-material voxel-based soft robots under the objective of locomotion.
This multi-material voxel-based approach is also used in another work by \textcite{Hiller2009} for a three-dimensional non-uniform beam, which is optimized to fit different deflected beam profiles or optimized to reach a maximum deflection area with the end-effector with two actuation forces applied along the length of the beam.
In a subsequent article, these voxel-based, non-uniform, three-dimensional beams have been additively manufactured and experimentally verified~\cite{Hiller2012}.

Recent advancements in the development of differentiable physical simulators, opened possibilities for gradient-based optimization in soft robotic design~\cite{Bacher2021, Cochevelou2023, Matthews2023, Gjoka2024, Bielawski2024}.
A requirement of these methods is a physical simulator that is written in a differentiable manner in each operation that is dependent on the design variables~\cite{Bacher2021}.
Another promising research direction in the design of soft pneumatic actuators is presented in \textcite{Chan2024}, where a latent diffusion model is used to design soft robots based on text prompts.

Current design optimization of soft pneumatic actuators are 
mostly concerned about the optimization of single chamber actuators
~\cite{Mosser2023, Zhang2017, Zhang2019c, Caasenbrood2020, Chen2019a, Matthews2023}, restricted to a handful of 
geometric properties~\cite{Runge2017b, Peng2019}, or tuned to reach a specific
target shape~\cite{Ma2017, Ding2019, Maloisel2021}.
Some design might not define a specific shape or physical interpretable
target~\cite{Chan2024}.

A detailed survey of automated design methods in soft robotics can be
found in \textcite{Pinskier2022}.
More specifically, for the application of black-box topology optimizations, where the objective function is possibly non-differentiable, a comprehensive overview is given in \textcite{Guirguis2020}.

The focus of the study presented here is on the design of soft pneumatic actuator through its two-dimensional circular cross-section.
However, we took inspiration by the voxel-based methods \cite{Hiller2009, Hiller2010, Hiller2010a, Hiller2012} and therefore rasterized the circular cross-section in a polar coordinate system.

The following sections of this article are structured in: \textit{Methods}, \textit{Results}, \textit{Discussion} and a final \textit{Conclusion \& Outlook} section.
In \autoref{sec:methods}, we cover the methodology required to perform the design optimization of the soft pneumatic actuators using genetic algorithms with geometrically exact beam models.
Three cases studies for different target workspaces were conducted, and the results are shown in \autoref{sec:results}.
These results are discussed in \autoref{sec:discussion}, followed by a conclusion of our study and a brief outlook on further research directions in \autoref{sec:conclusion_and_outlook}.

\section{Methods}
\label{sec:methods}

This section covers the necessary methods for performing a design optimization
of the cross-section of the soft pneumatic actuator using genetic algorithms.
In \autoref{sec:soft_pneumatic_actuator} the slender soft pneumatic actuator under consideration is presented.
The discretization of the cross-sectional design space, the geometrically exact beam model, and the description of the workspace for the pneumatic actuator are explained.
The genetic algorithm used in the black-box topology optimization to retrieve an optimal design for a given target workspace is described in \autoref{sec:genetic_algorithm}.

\subsection{Slender soft pneumatic actuators}
\label{sec:soft_pneumatic_actuator}

The initial shape of the slender soft pneumatic actuator is a cylinder with a ring cross-section, as shown in the left in \autoref{fig:disc}a).
The overall length of the actuator is set to $H = \qty{100}{\milli\meter}$ and does not include any caps to close off the air chambers.
These were neglected to simplify the model and make the results more interpretable, as without end caps no extension of the workspace that is dependent on the height of the end cap has to be considered.
Further, the inner radius of the ring is $r_s = \qty{10}{\milli\meter}$ while the outer radius is set to $r_e = \qty{25}{\milli\meter}$.
The length and radii of the soft pneumatic actuator are kept constant in this study.
This cylinder is then sculpted, sketched in \autoref{fig:disc}a), to include air chambers which can be pressurized such that an elongation and bending of the soft pneumatic actuator is achieved.
For this pressurization three different pressure supplies are available, and any number of air chambers can be assigned to a single pressure supply.
Except for a few special arrangements of air chambers, this results in a spatial deformation of the soft pneumatic actuator at different pressure combinations.
The sculpting of air chambers and assignment of pressure supplies takes place on the level of the discretized cross-section.

\subsubsection{Discretization of the cross-section}
\label{sec:discretization}

\begin{figure}
  \centering

  \textbf{a)}
  \includegraphics[width=0.19\textwidth, keepaspectratio]
    {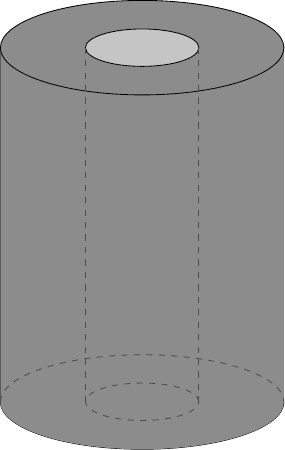}
  \hspace{0.01\textwidth}
  \largearrow
  \hspace{0.01\textwidth}
  \includegraphics[width=0.19\textwidth, keepaspectratio]
    {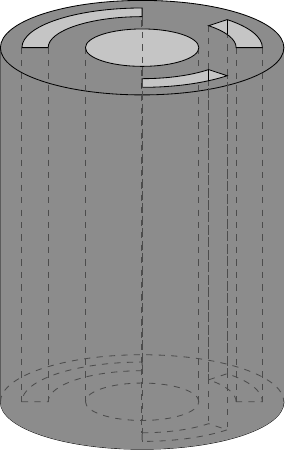}

  \vspace{1cm}

  \textbf{b)}
  \includegraphics[width=0.19\textwidth, keepaspectratio]
    {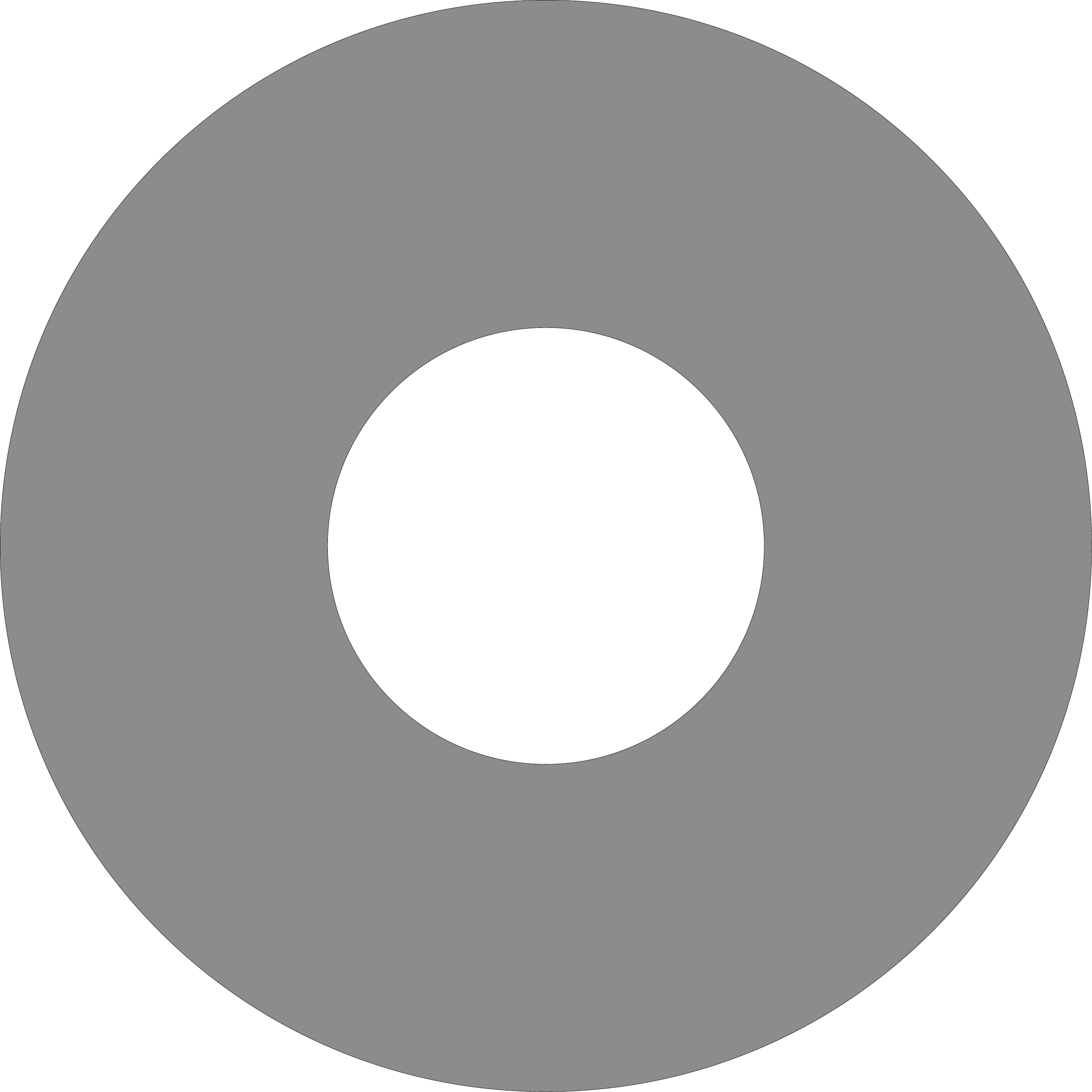}
    \hspace{0.01\textwidth}
    \smallarrow
    \hspace{0.01\textwidth}
  \includegraphics[width=0.19\textwidth, keepaspectratio]
    {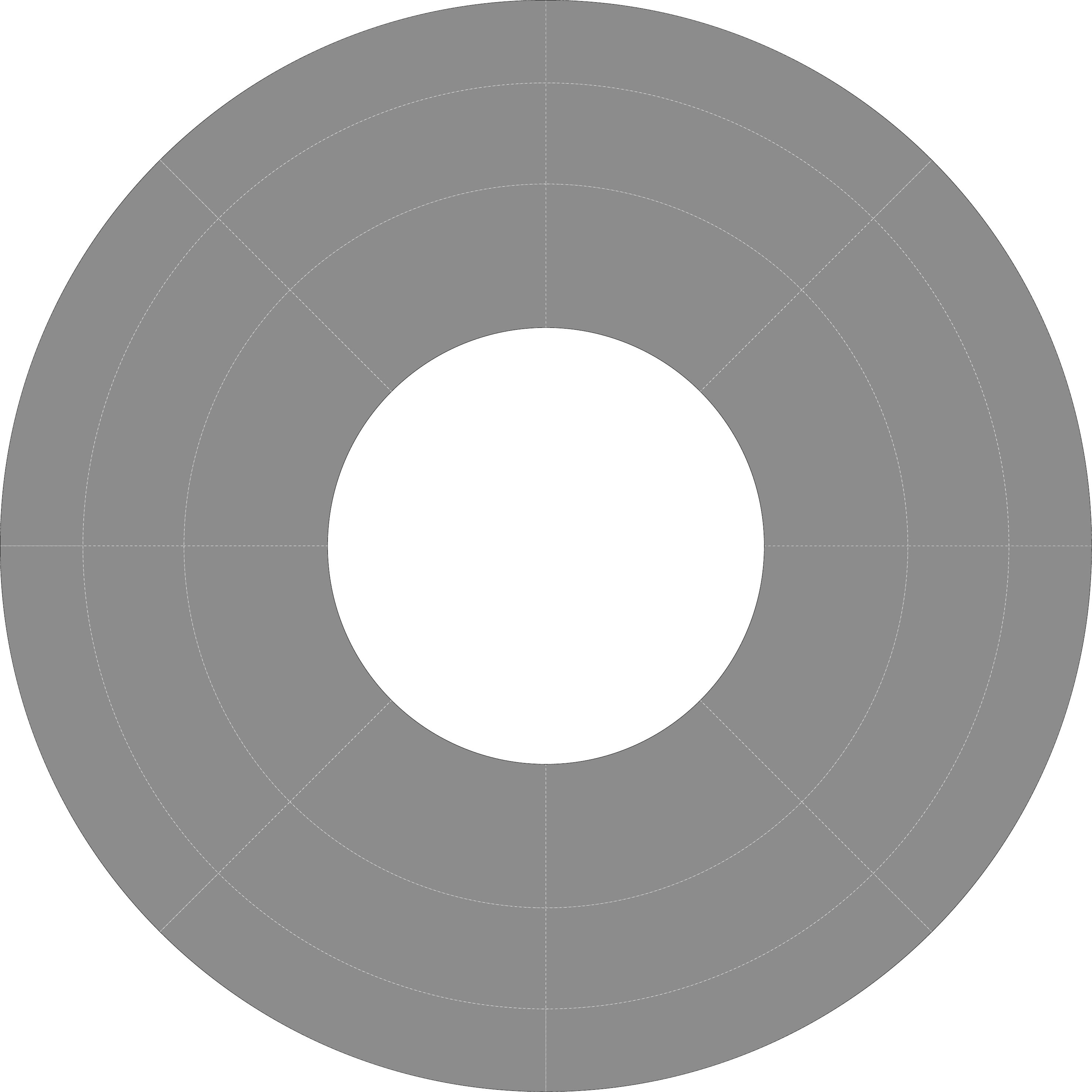}
    \hspace{0.01\textwidth}
    \smallarrow
    \hspace{0.01\textwidth}
  \includegraphics[width=0.19\textwidth, keepaspectratio]
    {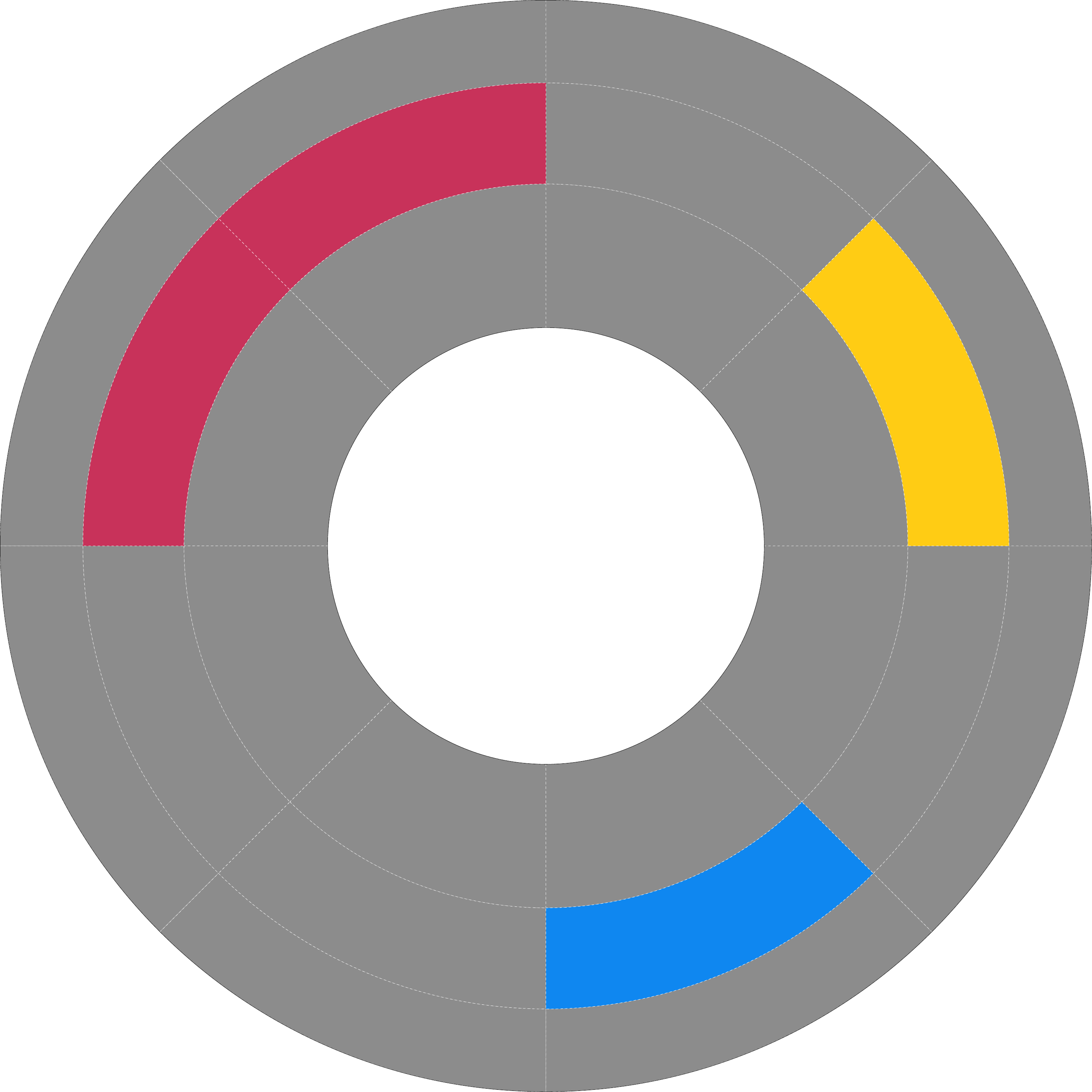}
  \caption{Discretization of the cross-section with \textbf{a)} the slender soft pneumatic actuator from initial cylindrical shape to a sculpted shape with air chambers and \textbf{b)} the discretization of the cross-section of this  cylindrical soft pneumatic actuator and assignment of air chambers (red, yellow, blue) and material (gray).}
  \label{fig:disc}
\end{figure}

The ring cross-section is discretized in a polar coordinate frame.
For this, the inner radius $r_s = \qty{10}{\milli\meter}$ and outer radius $r_e = \qty{25}{\milli\meter}$ are defined.
Furthermore, the possible angular values of the ring sector are confined by the angles $\varphi_s$ and $\varphi_e$.
For simplification, the starting angle is set to $\varphi_s = \qty{0}{\degree}$ and the end angle to $\varphi_e = \qty{360}{\degree}$, forming a full ring.
However, this method is not necessarily restricted to a whole ring and can be extended to ring sectors.

The ring is discretized with $N_r = 128$ elements in radial direction and  $N_\varphi = 360$ elements in circumferential direction.
The fidelity in circumferential direction is uniform, while in radial direction the fidelity decreases with greater radius such that the area of all elements have equal size.
This results in elements towards the outside of the ring spanning a smaller radial length.
In the step following the discretization the type of each element is set.
The element can either be filled with material or assigned to one of three possible  pressure supplies to form an air chamber.
We define the set of possible element types as $\mathcal{D} := \{ \text{Material}, \text{Chamber 1}, \text{Chamber 2}, \text{Chamber 3} \}$.
This discretization and assignment of pressure chambers is sketched in \autoref{fig:disc}b).

\subsubsection{Voronoi tessellation of the cross-sectional geometry}
\label{sec:genotype}

\begin{figure}
  \centering
  \includegraphics[keepaspectratio, width=0.3\linewidth]{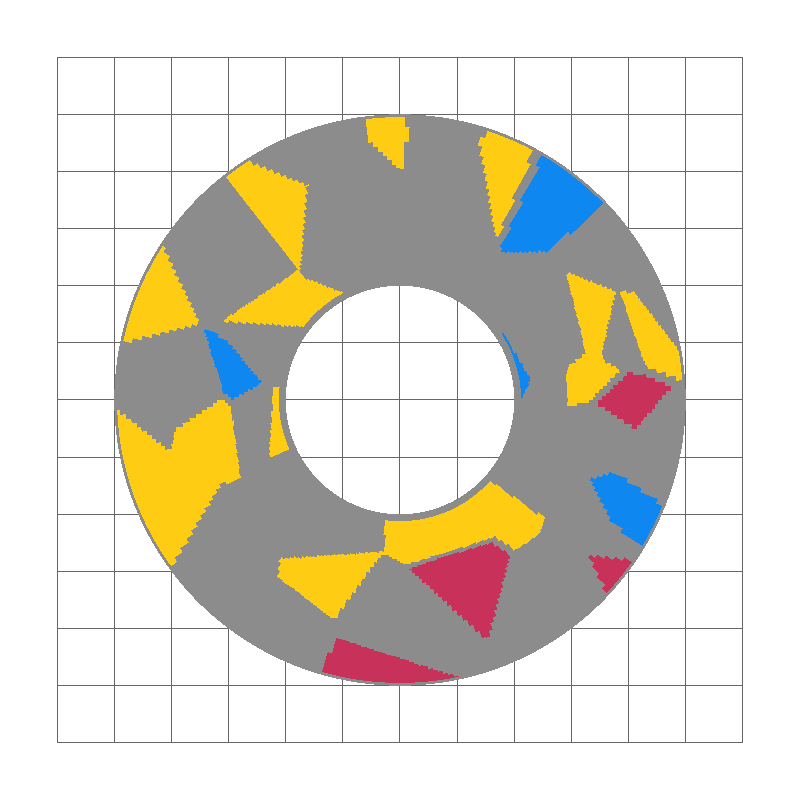}
  
  \caption{Discretized circular cross-section of a soft pneumatic actuator generated by a random Voronoi tessellation with air chambers (red, yellow, blue) and material (gray)}
  \label{fig:voronoi}
\end{figure}

The assignment of element types in the discretized cross-section is performed using Voronoi tessellations.
A detailed survey of Voronoi diagrams is given in \textcite{Aurenhammer1991}.
The algorithm for the generation of the cross-section is presented in Algorithm \ref{alg:voronoi}.
An arbitrary cross-section, filled with air chambers and material, is generated by randomly placing $N_f = 100$ feature points in the discretized cross-section.
Note that, due to the discretization in radial direction and the constraint of maintaining the same surface area for all elements, the feature points are not uniformly distributed in the cross-section and are more clustered towards the outer radius.
In addition to the position in polar coordinates, each feature point is randomly assigned a type from $\mathcal{D}$ that indicates whether this feature point corresponds to a site of material or site of a specific air chamber.
The probability of a feature point to be assigned to a material site is $p_m = 1/2$.
As we use pneumatic actuators with three different types of air chambers, each have an equal probability of $p_c = 1/6$ of occurring at a feature point.
This probability distribution ensures that there is enough material in the cross-section such that the soft robot is stiff enough to be stable under the applied pressure loads, but still provides enough space for air chambers.
After all feature points are randomly generated, the closest feature point to each discretized element is determined in Cartesian coordinates, and the element is assigned to the type of this feature point.
In an additional post-generation step, we ensure that a material wall exists between two different air chambers.
This is done by checking elements at the boundary of two sites with different types.
These elements are then assigned to material.
A wall thickness of four elements is used; however, due to the non-uniform discretization in radial direction, the wall thickness is not constant.
An exemplary cross-section is show in \autoref{fig:voronoi}a).
The first pressure chamber is shown in red, the second is shown in yellow, and the third is shown in blue.
Elements with material type are colored in gray.
It should be noted that there is no mechanism in place that restricts the generation of islands, meaning an air chamber is located within another air chamber.
In this case, not all elements with material type are necessarily connected to one another.

\begin{algorithm}
    \caption{Voronoi tessellation of the cross-sectional geometry}\label{alg:voronoi}
  \begin{algorithmic}
  \State \textbf{Initialization}
  \State $N_r \gets 128$ \Comment{Number of elements in radial direction}
  \State $N_\varphi \gets 360$ \Comment{Number of elements in circumferential direction}
  \State $N_f \gets 100$ \Comment{Number of feature points}
  \State $\mathcal{D} \gets \{ \text{Material}, \text{Chamber 1}, \text{Chamber 2}, \text{Chamber 3} \}$ \Comment{Set of possible element types}
  \State $p_m \gets 1/2$ \Comment{Probability of material}
  \State $p_c \gets 1/6$ \Comment{Probability of chamber}
    \\
  \State \textbf{Feature point generation}
  \For{$i \gets 1,\ N_f$}
    \State $n_r \gets \text{random}(\{ z \in \mathbb{Z}\ |\ z \le N_r \})$ \Comment{Element index in radial direction}
    \State $n_\varphi \gets \text{random}(\{ z \in \mathbb{Z}\ |\ z \le N_\varphi \})$ \Comment{Element index in circumference direction}
    \State $t \gets \text{random}(\mathcal{D};\ ( p_m,\ p_c,\ p_c,\ p_c ) )$ \Comment{Type of feature point}
    \State $f_i \gets (r,\ \varphi,\ t)$
  \EndFor
    \\
  \State \textbf{Type assignment on element level}
  \For{$i \gets 1,\ N_r$}
    \For{$j \gets 1,\ N_\varphi$}
      \State $d \gets \infty$
      \For{$k \gets 1,\ N_f$}
        \State $d' \gets \text{distance}(f_k,\ (i,\ j))$ \Comment{Distance is computed in Cartesian coordinates}
        \If{$d' < d$}
          \State $d \gets d'$
          \State $t \gets f_k[2]$
        \EndIf
      \EndFor
      \State $c_{i,j} \gets t$ \Comment{Assign type to element}
    \EndFor
  \EndFor

  \end{algorithmic}
\end{algorithm}

\subsubsection{Global and local coordinate systems}

The fixed end of the slender soft pneumatic actuator is set to coincidence with the origin of the global coordinate system.
Furthermore, the undeformed backbone is aligned at the upright coordinate axis.
The target workspaces, which are specified later, are defined in the global coordinate systems and include this undeformed state.
This description generalizes, as the target workspace can be transformed into this reference frame.

The orientation of the cross-section is shifted and rotated to move the center of mass to the origin of the global coordinate system.
The simulations are performed in the coordinate frame defined by the principal axes of inertia, which simplifies the simulation of the backbone, as deviatoric moments of inertia can be neglected and the undeformed backbone in the unpressurized state is always aligned with the upright axis of the global coordinate system.
The transformation between the principal axes coordinate system and the global coordinate system is described by a rotation around the upright axis.

\subsubsection{Geometrically exact beam models}
\label{sec:beam_models}

To model the behavior of the slender soft pneumatic actuator, a one-dimensional geometrically exact beam model, specifically the Cosserat beam model, is used.
This geometrically exact theory is necessary as the soft pneumatic actuator is subject to large deformations and rotations when pressurized due to its soft material and flexible structure.
The following description of the Cosserat beam theory is based on the detailed explanation in \textcite{Till2019}.
The pressurization of the soft pneumatic actuator is assumed to be quasi-static.
Therefore, the dynamic terms are simplified to the static case in the formulation of the beam model by neglecting all time-derivates ${\partial\left(\bullet\right)}/{\partial t} = 0$.

The centerline of the beam is described by the position $\mathbf{h}(s)$, depending on the undeformed arc length $s$ in the global coordinate frame $\{ \mathbf{e}_1, \mathbf{e}_2, \mathbf{e}_3 \}$, as follows
\begin{equation}
  \mathbf{h}(s) = x(s) \mathbf{e}_1 + y(s) \mathbf{e}_2 + (z(s) + s) \mathbf{e}_3\ ,
\end{equation}
where $x(s)$, $y(s)$ and $z(s)$ are the displacements of points along the centerline of the beam in global coordinates.
The strains $\mathbf{v}(s)$ from elongation and shearing can be derived from the deformation of the centerline using
\begin{equation}
  \mathbf{v}(s) = \mathbf{R}(s) \frac{\partial \mathbf{h}(s)}{\partial s}
\end{equation}
where $\mathbf{R}(s) \in SO(3)$ is a rotation matrix describing the transformation 
from the global to the local coordinate frame.
The rotation matrix is formulated in terms of quaternionic orientation.
With the rotation matrix, it is possible to compute the curvature $\mathbf{u}(s)$ due to 
bending and torsion
\begin{equation}
  [u]_\times(s) = \mathbf{R}^T(s)\frac{\partial\mathbf{R}(s)}{\partial s}
\end{equation}
where $[\bullet]_\times$ denotes the skew-symmetric matrix for the cross product.

The material behavior is assumed to be isotropic and linear-elastic.
Therefore, the contact force $\mathbf{n}$ can be specified by
\begin{equation}
  \mathbf{n}(s) = \mathbf{R}(s) \mathbf{K} (\mathbf{v}(s) - \mathbf{v^*})\ , \label{eq:material_n}
\end{equation}
the contact torque $\mathbf{m}$ can be calculated using
\begin{equation}
  \mathbf{m}(s) = \mathbf{R} \mathbf{J} (\mathbf{u}(s) - \mathbf{u^*})\ .\label{eq:material_m}
\end{equation}
Both $\mathbf{K}$ and $\mathbf{J}$ are stiffness matrices describing the isotropic linear-elastic material law.
Precisely, $\mathbf{K}$ describes elongation and shear stiffness, while $\mathbf{J}$ describes torsional and bending stiffness.
Considering a beam that is already deformed in its stress-free state, $\mathbf{v^*}$ and $\mathbf{u^*}$ describe the strains and curvature in this state respectively.
In the case of a simple initial straight beam, as considered here, $\mathbf{v^*} = (0,0,1)^T$ and $\mathbf{u^*} = (0, 0, 0)^T$.
Using the Young`s modulus $E$ and the shear modulus $G$ the material matrices can be written as 
\begin{equation}
  \mathbf{K} = \begin{bmatrix}
    \kappa GA & 0 & 0 \\
    0 & \kappa GA & 0 \\
    0 & 0 & EA
  \end{bmatrix}
  \ ,
\end{equation}
and
\begin{equation}
  \mathbf{J} = \begin{bmatrix}
    EI_1 & 0 & 0 \\
    0 & EI_2 & 0 \\
    0 & 0 & EI_T
  \end{bmatrix}
  \ ,
\end{equation}
assuming the local coordinate system to align with principal axes of inertia in the center of mass.
For a linear-elastic material law, these matrices contain the material properties $E$ and $G$ as well as the geometric properties of the cross-section.
These geometric properties are the area moment of inertia around the first principal axis $I_1$, the area moment of inertia around the second principal axis $I_2$, the torsional moment of inertia $I_T$ and the shear correction factor $\kappa$.

The equilibrium state of the beam is described with 
\begin{equation}
  \frac{\partial \mathbf{n}(s)}{\partial s} = -\mathbf{f}_e(s) -\mathbf{f}_p(s)\ ,
\end{equation}
\begin{equation}
\frac{\partial \mathbf{m}(s)}{\partial s} = 
-\frac{\partial \mathbf{h}(s)}{\partial s} \times \mathbf{n}(s) - \mathbf{l}_e(s) - \mathbf{l}_p(s)
\end{equation}
where $\mathbf{f}_e(s)$ and $\mathbf{f}_p(s)$ are distributed force loads and $\mathbf{l}_e(s)$ and $\mathbf{l}_p(s)$ are distributed torque loads.
More precisely, $\mathbf{f}_e$ and $\mathbf{l}_e$ are distributed external loads, and $\mathbf{f}_p(s)$ and $\mathbf{l}_p(s)$ are distributed loads due to the pressurization of the soft actuator, which is described in the following section.

In this study, the Young's modulus is set to $E = \qty{300}{\kilo\pascal}$ and the shear modulus to $G = \qty{100}{\kilo\pascal}$ to model an incompressible material behavior.
This Young's modulus is in the stiffness range of silicones~\cite{Xavier2021}.
Furthermore, torsional effects ($I_{T} = 0$) and the shear correction ($\kappa = 1$) were neglected.
The soft pneumatic actuator is fixed on the end at $s = 0$ and is free on the other end at $s = L$.
The influence of gravity and other forces, besides the pressure forces, is neglected.
Therefore, the distributed external loads in this study are $\mathbf{f}_e = 0$ and $\mathbf{l}_e = 0$.
Although an incompressible material behavior is assumed, no necking of the slender actuator is considered, and the cross-sectional area stays constant in our model.
Contact forces due to necking, under the assumption of incompressible material behavior, are therefore neglected.

\subsubsection{Pressure loads}
\label{sec:pressure_loads}

In extension to the Cosserat beam model described in the previous section, 
pressure loads are applied to the model of the soft pneumatic actuator.
These pressure loads are derived in \textcite{Till2019} and described in the following for the sake of completeness.
The air pressure, used to actuate the soft robot, results in distributed loads along the length of the actuator and in point loads at the boundaries, which must be considered in the simulation.
Since in this study the cross-section of the soft pneumatic actuator is based on a polar grid, the pressure loads are computed based on the summation of the grid elements.

A single grid element has a constant cross-sectional area $\Delta A$.
The area $A_i = N_i \Delta A$ defines the area for all grid elements that belong to the same pressure supply, where $N_i$ is the number of grid elements for that pressure supply.
In addition to the cross-sectional area $A_i$ the point of attack of the pressure force is described by $\mathbf{r}_i$ and can be computed analogous to the center of mass of grid elements filled with material.
This point of attack is formulated in relation to the center of mass in the principal axis coordinate system.
The force point load and torque point load due to the pressurization of all chambers for one pressure supply are
\begin{equation}
  \mathbf{F}_i = P_i A_i \mathbf{e}_3\ ,
\end{equation}
\begin{equation}
  \mathbf{L}_i = \mathbf{r}_i \times \mathbf{F}_i
\end{equation}
accordingly, where $P_i$ is the pressure applied by a single pressure supply.
The point loads are applied to the ends $s = 0$ and $s = L$ of the beam were the air chambers are assumed to be closed off.
In addition to these point loads, distributed loads must be taken into account.
These distributed loads act due the bending and the pressurization of the soft pneumatic actuator
\begin{equation}
  \mathbf{f}_p(s) = -\sum\limits_{i=1}^{3} P_i A_i \frac{\partial\mathbf{R}(s)}{\partial s} \ \mathbf{e}_3\ ,
\end{equation}
\begin{equation}
  \mathbf{l}_p(s) = -\sum\limits_{i=1}^{3} P_i A_i \mathbf{R}(s) \left[
     \left(\mathbf{v}(s) + \mathbf{u}(s) \times \mathbf{r}_i \right) \times \mathbf{e}_3 
     + \mathbf{r}_i \times (\mathbf{u}(s) \times \mathbf{e}_3)
  \right]\ .
\end{equation}

\subsubsection{Area moments of inertia}

The area moments of inertia $I_{1}$ and $I_{2}$ are calculated on the basis of the discretized cross-sectional geometry.
For each discretized element, the area moments of inertia are computed relative to the center point of the entire ring geometry and then summed up to obtain the area moments of inertia for the entire structure.
After the computation of the area moments of inertia regarding the center point of the entire ring geometry, the area moments of inertia are transformed to the center of mass of the cross-section using the Steiner theorem.

\subsubsection{Workspace description}
\label{sec:workspace}

\begin{figure}
  \centering

  \textbf{a)}
  \includegraphics[keepaspectratio, width=0.3\linewidth]{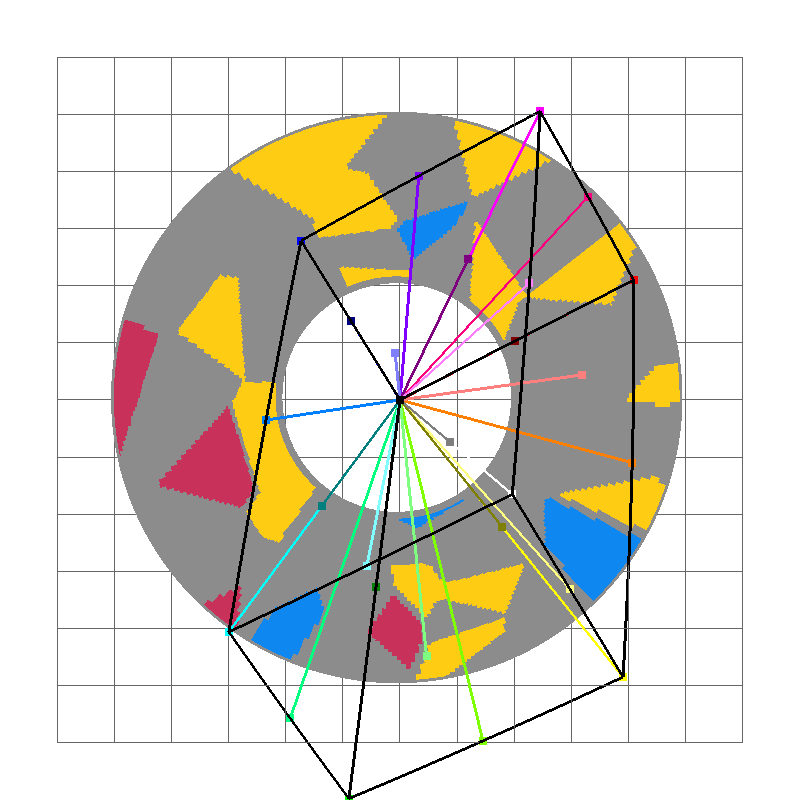}
  \textbf{b)}
  \includegraphics[keepaspectratio, width=0.3\linewidth
  , trim={3cm 4cm 3cm 2cm}, clip]{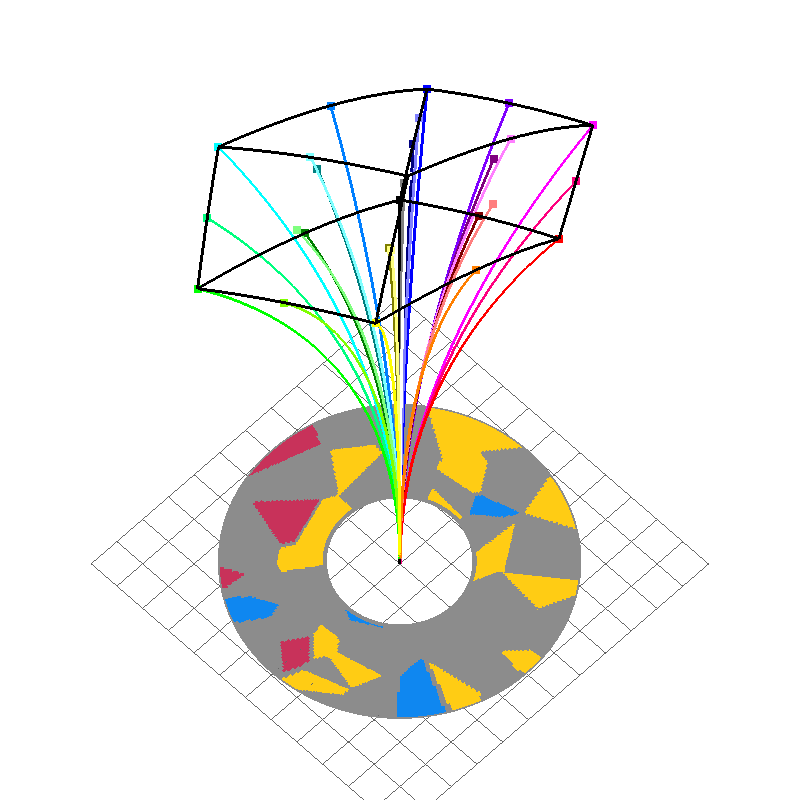}
  
  \caption{\textbf{a)} Randomized circular cross-section with the corresponding workspace of the end-effector rendering in top-down view, and \textbf{b)} the same cross-section and workspace rendered in a diagonal view.}
  \label{fig:workspace}
\end{figure}

The workspace of the actuator is described by the spatial positions of the end-effector at different actuation parameters.
In the case of a slender soft pneumatic actuator, these parameters are the air pressures in the individual air chambers.
The use of three different pressure supplies cause the end-effector of the actuator to move in a three-dimensional space.
By setting the maximal pressure $P_\text{max}$, which the pressure supply can provide, or the soft pneumatic actuator can withstand without rupture, this three-dimensional space is a closed domain in $\mathbb{R}^3$.
This closed domain is the workspace $\mathcal{W} \in \mathbb{R}^3$ 
of the actuator, described by the possible positions of the actuator's end-effector.

Due to the highly non-linear mapping between the input pressures and the position of the end-effector, the workspaces cannot be described by linear geometric relations.
However, a good approximation of this workspace is provided by quadratic deformed hexahedra, similar to the formulation of quadratic elements in the finite element method.
Using this approximation, the workspace is characterized by 27 different end-effector positions at different pressure combinations.
The maximum allowed pressure is set to $P_\text{max} = \qty{100}{\kilo\pascal}$.
With this, the 27 end-effector positions can be obtained by computing the backbone using a set of pressures for each possible combination of actuation pressure $P_i \in \{\qty{0}{\kilo\pascal}{,}\, \qty{50}{\kilo\pascal}{,}\, \qty{100}{\kilo\pascal}\}$.

For the cross-sectional geometry given in \autoref{fig:voronoi}, the workspace is overlaid in \autoref{fig:workspace}a) in a top-down view and in \autoref{fig:workspace}b) in diagonal view.
In addition to the workspace, the backbones of the deformed actuator, leading to the workspace description, are rendered.
Note that these backbones are colored based on the pressure influence of each chamber, with colors representing a combination of \textit{red} (first chamber), \textit{green } (second chamber), \textit{blue} (third chamber) in the RGB color format.
This color coding slightly deviates from the color schema used in the rendering of the cross-sections to allow for a direct mapping between pressure intensity and byte sized color intensity.
In subsequent simulations, the backbones are discarded after computing the end-effector positions, as only the latter are needed to describe the workspaces.

\subsection{Genetic Algorithm}
\label{sec:genetic_algorithm}

We use a genetic algorithm to perform the black-box topology optimization of the cross-sectional geometry of the slender soft pneumatic actuator.
Genetic algorithms are gradient-free, evolutionary optimization algorithms that mimic the evolution of lifeforms, where a large population of individuals is simulated simultaneously and ranked according to their fitness~\cite{Kane1996}.
Individuals that perform better than others are kept for subsequent generations, while poorly performing individuals are discarded~\cite{Kane1996}.
Between generations, some individuals can produce new combinations of individuals in a recombination process, other individuals are randomly altered through mutation methods~\cite{Kane1996}.

The genetic algorithm used in this study is an adaptation of the presented genetic algorithm in \textcite{Pernkopf2005}.
Therefore, we refer to this as detailed source for the following explanations in this section.
In the first step, the initial population is generated.
The number of individuals in each population stays constant at $N_p = 20$ individuals.
The genotype of an individual is based on the feature points from the Voronoi tessellation and the phenotype is the resulting reachable target workspace.
The genetic algorithm is then performed over $N_g = 100$ generations, without any additional convergence criteria being used.
In each generation three steps are performed: \textbf{recombination} to generate new offspring, \textbf{selection} of the best individuals, and \textbf{mutation}.
We tested two recombination methods based on random crossover and two mutation methods.
These are detailed in the following sections.

\subsubsection{Recombination}

In the recombination process new offspring are generated.
The number of offspring individuals $N_o = p_c N_p$ is based on the crossover probability $p_c = 0.8$ and the population size $N_p$.
Two offspring are formed by the crossover of two parent individuals, which are randomly selected.
We employed two different strategies for the recombination of the parents to generate new offspring.

\begin{itemize}
  \item \textbf{Range-based crossover (range)} \\
  The range-based crossover is inspired by the 2-point crossover in~\cite{Kane1996}.
  Two angles $\varphi_a$ and $\varphi_b$ are randomly selected in each procedure to generate two offspring.
  All feature points in the interval $[\varphi_a, \varphi_b]$ are swapped between parents to form two new offspring.
  It must be emphasized that this method does not keep the number of feature points constant between individuals, as there can be a different number of feature points in the randomly selected interval.

  \item \textbf{Crossover keeping the amount of feature points constant (fixed)} \\
  This crossover method is referred to as fixed amount crossover and is designed to keep the amount of swapped feature points constant.
  Therefore, the overall count of feature points in each individual remains constant.
  We used a ratio of $\nu = 0.5$ meaning that we swap half the feature points.
  In our case this is equal to $50$ feature points.
  For the swapping, a random angle is chosen for each crossover combination, and the nearest feature points, with the smallest angular distance, are swapped between the parents, generating two new offspring.
\end{itemize}

\subsubsection{Selection}

As the number of individuals between each generation remains constant, and there are now parents and offspring as possible candidates for the next generation, only the individuals with the highest performance are selected.
All other individuals are discarded to form the population for the next generation.
Instead of selecting individuals with high fitness values, as in~\cite{Kane1996}, best-performing individuals are selected based on the total loss of the individual.
The total loss $T$ is computed with
\begin{equation}
  T = L + K,
\end{equation}
where $L$ is the loss due to the deviation to the given target workspace as first objective, and $K$ is the loss from the secondary objective.
In this study these two objectives are considered for the optimized designs.
The former one is the primary objective that needs to be fulfilled as closely as possible, while the latter one is a secondary objective that can be relaxed.
Both are linearly scalarized into the single objective function $T$.

\paragraph{First Objective}

The first objective describes the approximation of the target workspace.
The loss is defined as the sum of squared distances of the computed workspace to the target workspace.
Numerical computations are performed in \unit{\milli\meter}, the default unit for distance in our simulations.
For readability, we omit this unit in the computation of the loss.
Thus, the loss can be computed as 
\begin{equation}
  L = \sum\limits_{j=0}^{27} \left[
    \mathbf{h}(H,\ P_{1,j},\ P_{2,j},\ P_{3,j}) - \mathbf{w}_j\right]^2\ ,
\end{equation}
where $\mathbf{w}_j$ is the $j$-th node defining the target workspace at the pressure combinations $(P_{1,j},\ P_{2,j},\ P_{3,j})$.

\paragraph{Second Objective}

The second objective is used to focus the optimization on the formulation of clustered sites.
The loss term is applied to penalize the formation of many clusters of chambers.
One possible solution for this penalty is to count the number of individual chambers.
However, this would require a knowledge of the connected feature points.
This connection of feature points is computed using the Delaunay triangulation~\cite{Aurenhammer1991}.
Since this triangulation yields the edges of connected feature points, an alternative approach to counting the number of chambers is penalizing the ratio of edges that connect different feature points.
By using this ratio the loss term is indifferent to the overall number of simplices computed from the triangulation, regardless of the placement of feature points.
The loss term $K$ is formulated as
\begin{equation}
  K = \lambda r
\end{equation}
where $\lambda$ is the weighing factor in relation to the first objective and $r$ is the ratio of edges with different types of feature points to edges with the same type of feature points.
We employ a fixed factor of $\lambda = \qty{1000}{}$.

\subsubsection{Mutation}
After selecting the individuals for the next generation, a subset of these individuals is selected for mutation.
The ratio of individuals that are mutated is defined by $p_m = 0.1$.

The idea behind the mutation method described here is similar to the boundary mutation method described in~\cite{Kane1996}.
We employ the Delaunay triangulation to compute a graph of connected feature points, referred to as edges in the following.
Edges where one feature point has a different material or chamber assignment than the other, must describe a boundary of a larger region.
In the mutation process, edges with these properties are randomly selected for mutation.
The number of selected edges is determined by the ratio $p_f = 0.1$ and the number of feature points used.
For each of these edges one of the two feature points is randomly selected for mutation and two different methods have been tested for this boundary mutation.

\begin{itemize}
  \item \textbf{Direct mutation} (direct) \\
  The direct mutation method copies the value of the other feature point
  on the selected edge.

  \item \textbf{Weighted mutation} (weighted) \\
  The weighted mutation method considers all neighbor feature points 
  connected to the selected feature point.
  The probability of the new type is based on the number of neighbor feature points from different types.
  Thereby, neighbor feature points of the same type as the selected one are not counted.
  This ensures that the value of the selected feature point must change, and the mutation is performed regardless.

\end{itemize}

\section{Results}
\label{sec:results}

\begin{figure}[h!]  
  \centering

    \phantom{\textbf{a)}}
    \includegraphics[keepaspectratio, width=0.30\textwidth]
      {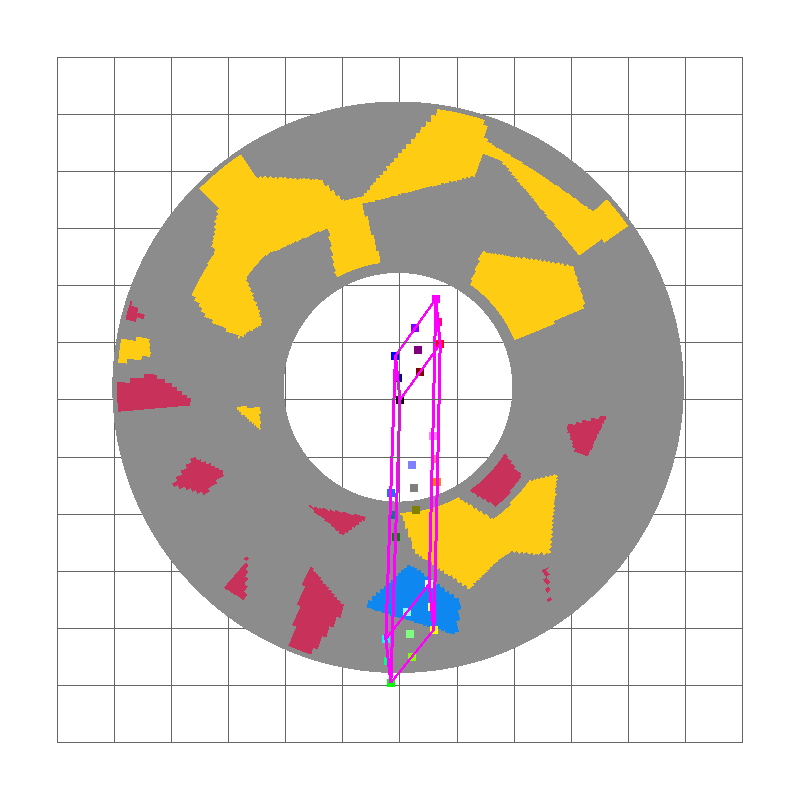}
    \phantom{\textbf{b)}}
    \includegraphics[keepaspectratio, width=0.3\textwidth]
      {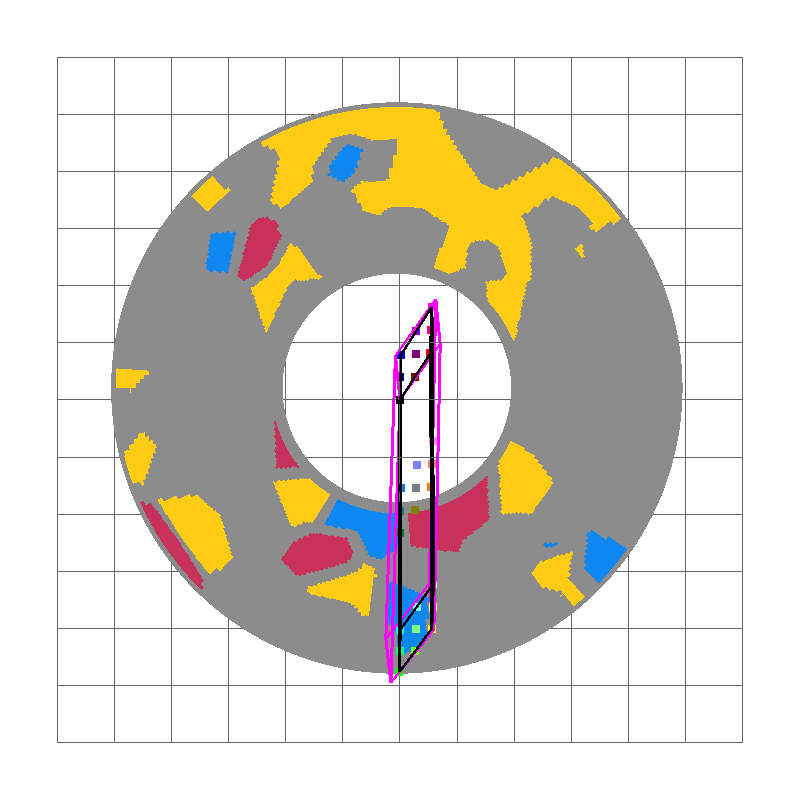}
    \phantom{\textbf{c)}}
    \includegraphics[keepaspectratio, width=0.3\textwidth]
      {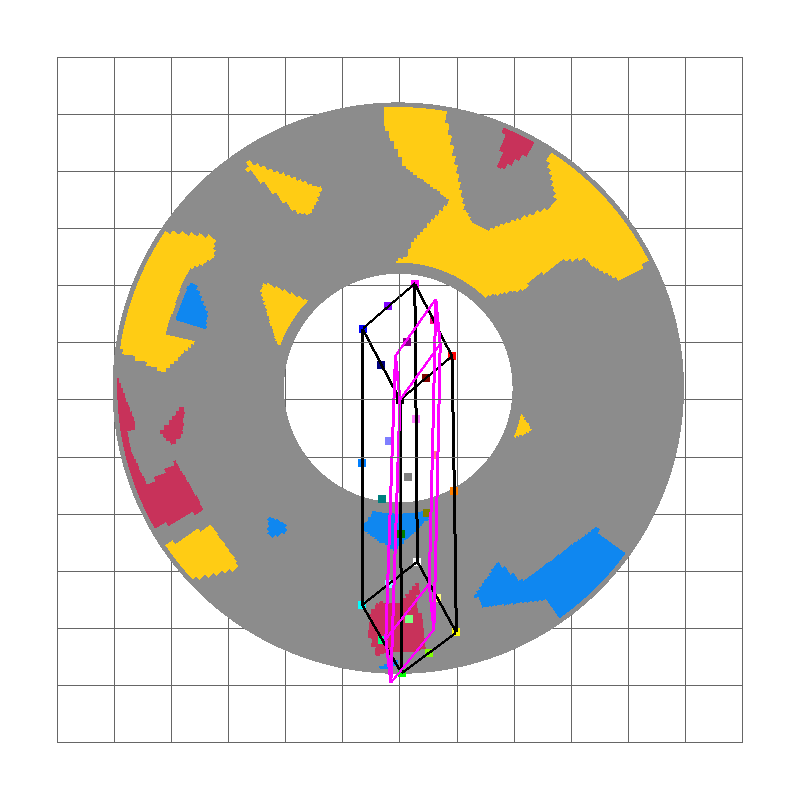}
    \\
    \textbf{a)}
    \includegraphics[keepaspectratio, width=0.3\textwidth,
      trim={3cm 4cm 3cm 3cm}, clip]
      {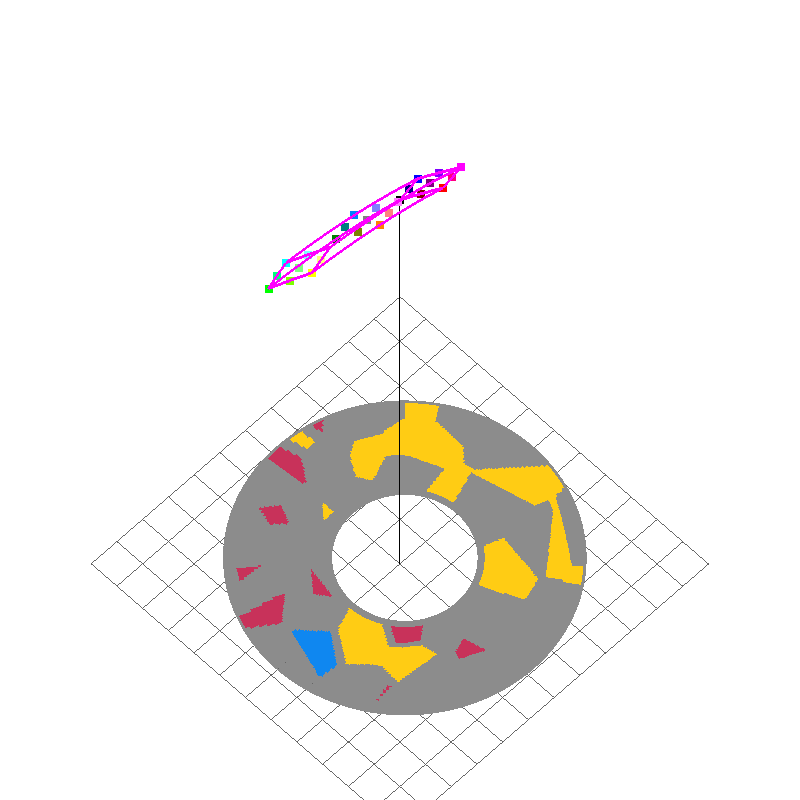}
    \textbf{b)}
    \includegraphics[keepaspectratio, width=0.3\textwidth,
    trim={3cm 4cm 3cm 3cm}, clip]
      {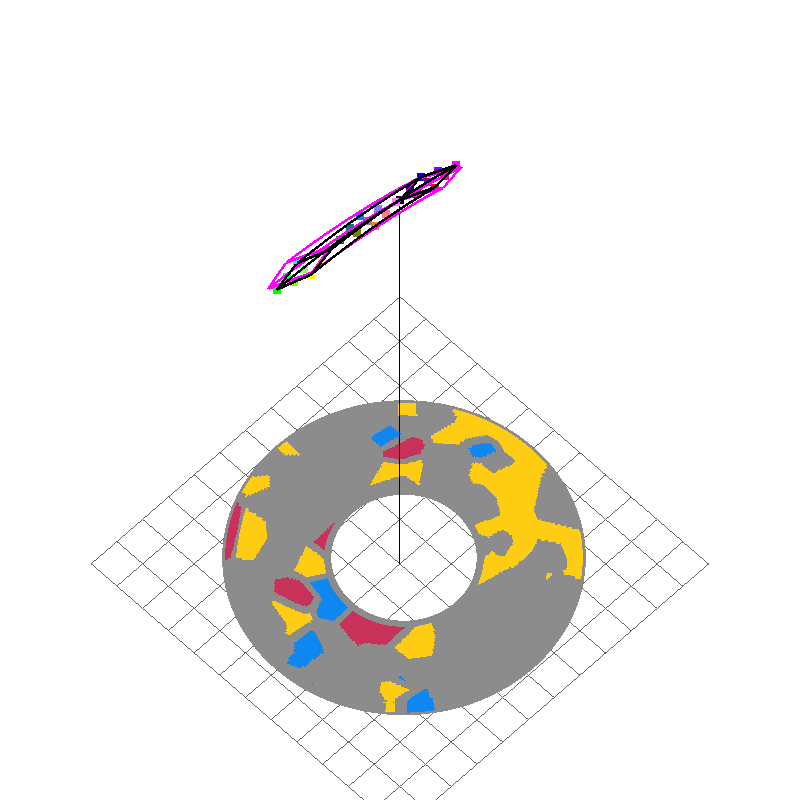}
    \textbf{c)}
    \includegraphics[keepaspectratio, width=0.3\textwidth,
    trim={3cm 4cm 3cm 3cm}, clip]
      {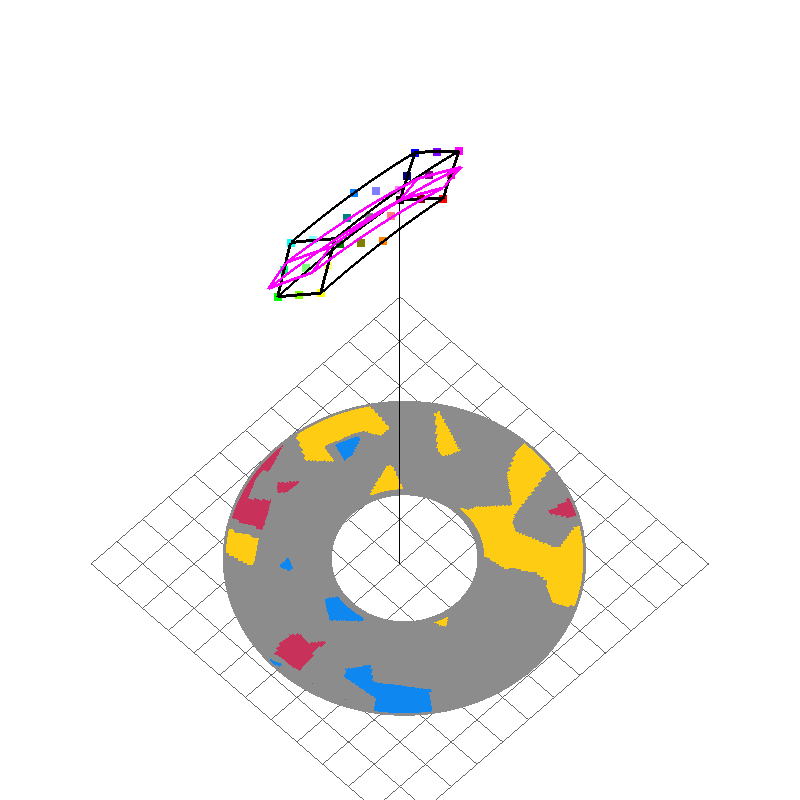}
      
    \caption{
      First case study with a random cross-section to generate target workspace.
      \textbf{a)} The randomly generated cross-section and the computed workspace as target workspace in top-down view and diagonal view,
      \textbf{b)} the best individual (cross-section and workspace) obtained through all iterations and methods and the target workspace (pink),
      \textbf{c)} the best individual (cross-section and workspace) obtained at the end of the worst performing iteration through all iterations and methods and the target workspace (pink)}
    \label{fig:optimized_design_a}
\end{figure}

\begin{figure}[h!]
    \centering
      \phantom{\textbf{a)}}
      \includegraphics[keepaspectratio, width=0.30\textwidth]
        {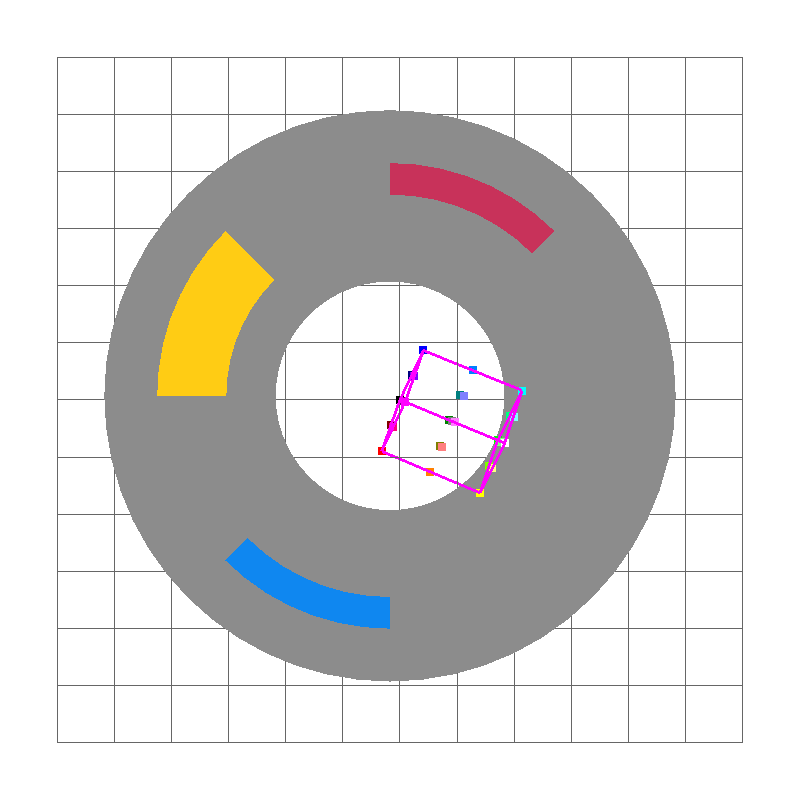}
      \phantom{\textbf{b)}}
      \includegraphics[keepaspectratio, width=0.3\textwidth]
        {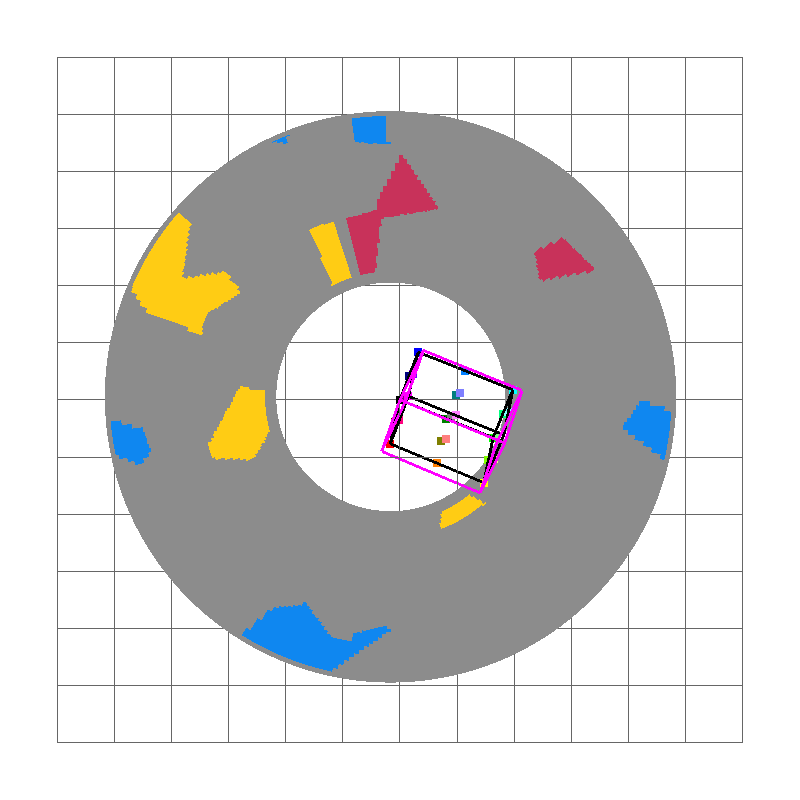}
      \phantom{\textbf{c)}}
      \includegraphics[keepaspectratio, width=0.3\textwidth]
        {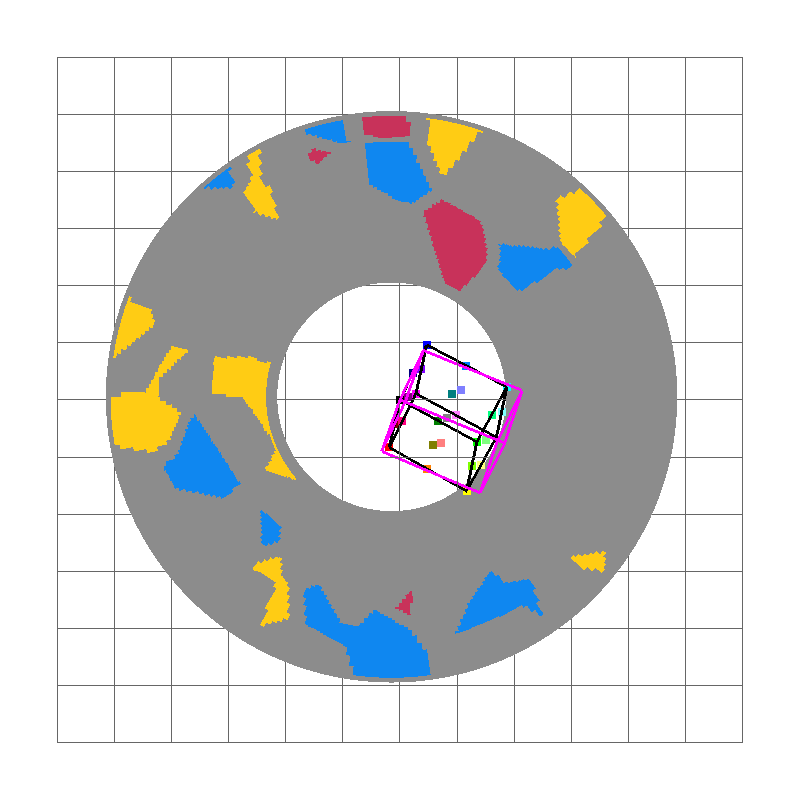}
      \\
      \textbf{a)}
      \includegraphics[keepaspectratio, width=0.3\textwidth,
      trim={3cm 4cm 3cm 4cm}, clip]
        {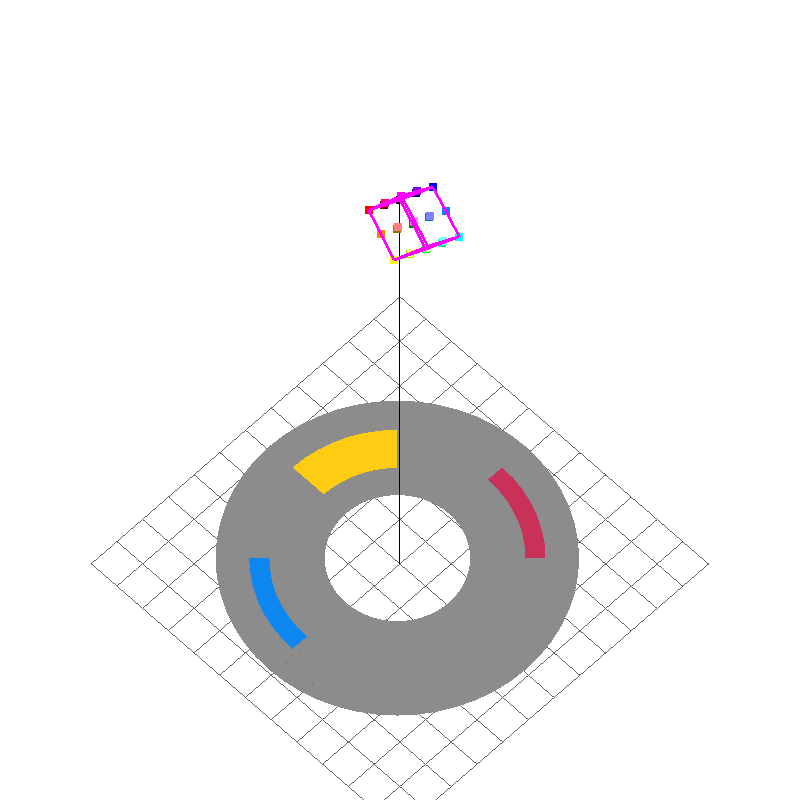}
      \textbf{b)}
      \includegraphics[keepaspectratio, width=0.3\textwidth,
      trim={3cm 4cm 3cm 4cm}, clip]
        {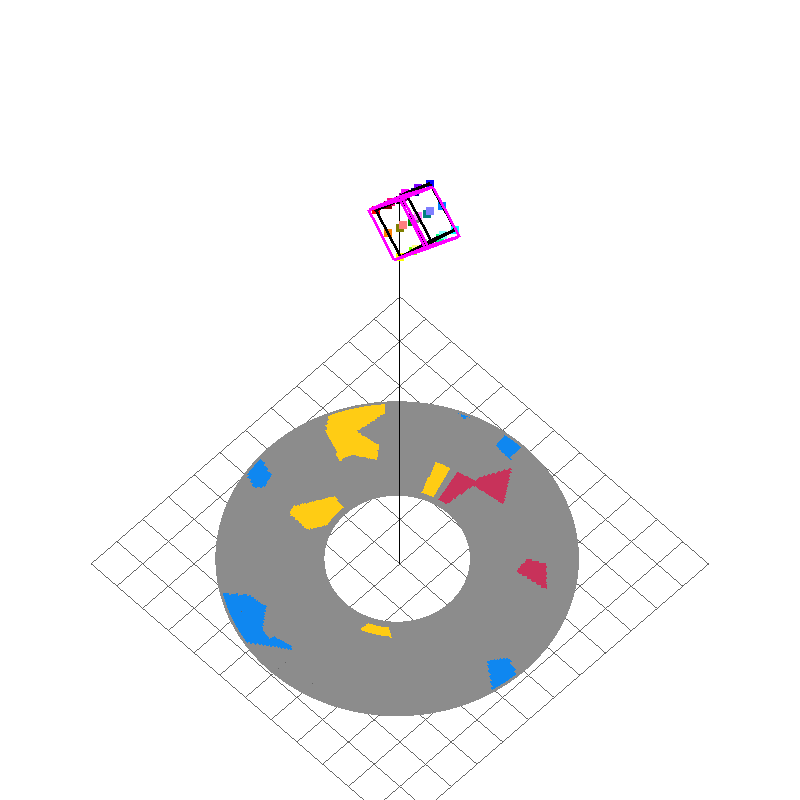}
      \textbf{c)}
      \includegraphics[keepaspectratio, width=0.3\textwidth,
      trim={3cm 4cm 3cm 4cm}, clip]
        {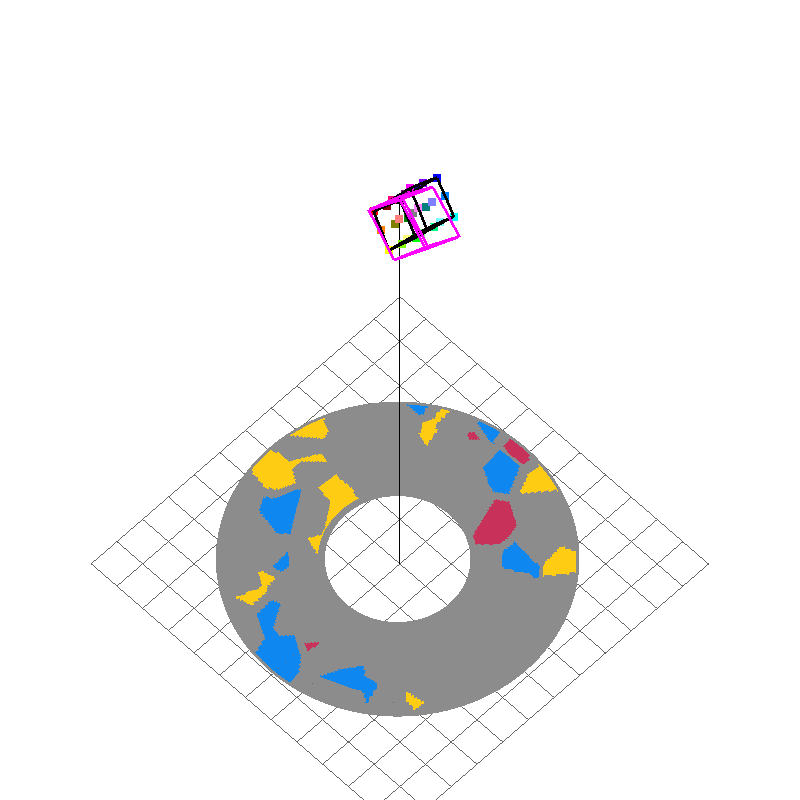}
        
      \caption{
        Second case study with a specific cross-section to generate target workspace.
        \textbf{a)} The manually designed cross-section and the computed workspace as target workspace in top-down view and diagonal view,
        \textbf{b)} the best individual (cross-section and workspace) obtained through all iterations and methods and the target workspace (pink),
        \textbf{c)} the best individual (cross-section and workspace) obtained at the end of the worst performing iteration through all iterations and methods and the target workspace (pink)}
      \label{fig:optimized_design_b}
\end{figure}

\begin{figure}[h!]
  \centering
      \phantom{\textbf{a)}}
      \includegraphics[keepaspectratio, width=0.30\textwidth]
        {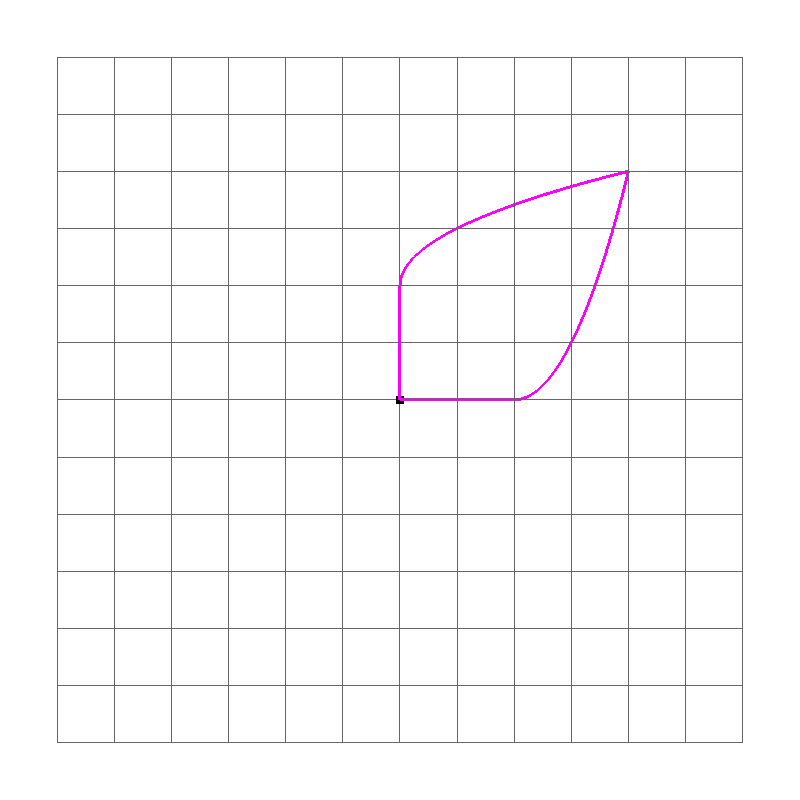}
      \phantom{\textbf{b)}}
      \includegraphics[keepaspectratio, width=0.3\textwidth]
        {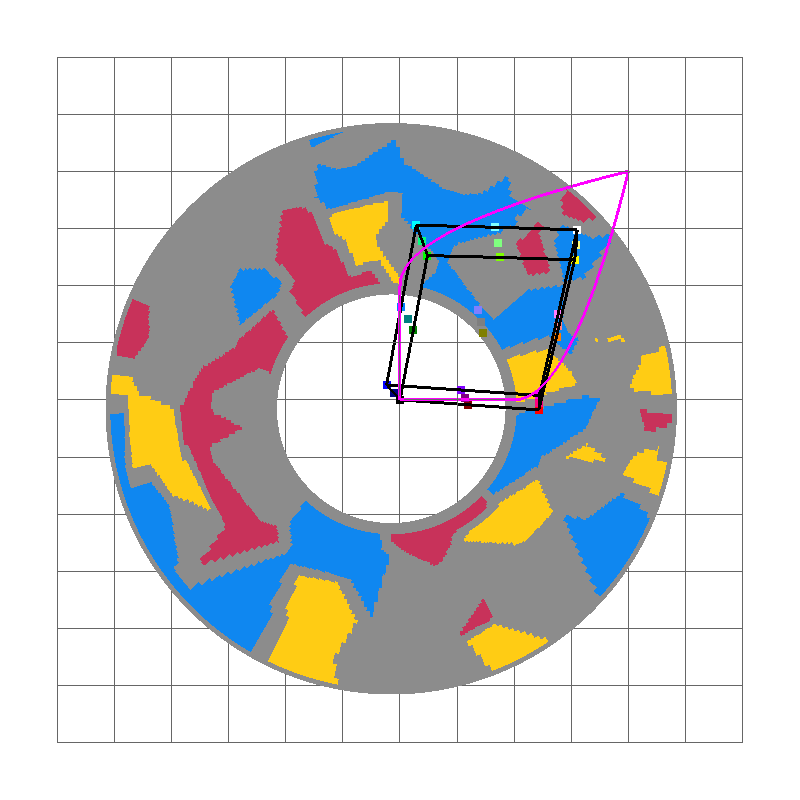}
      \phantom{\textbf{c)}}
      \includegraphics[keepaspectratio, width=0.3\textwidth]
        {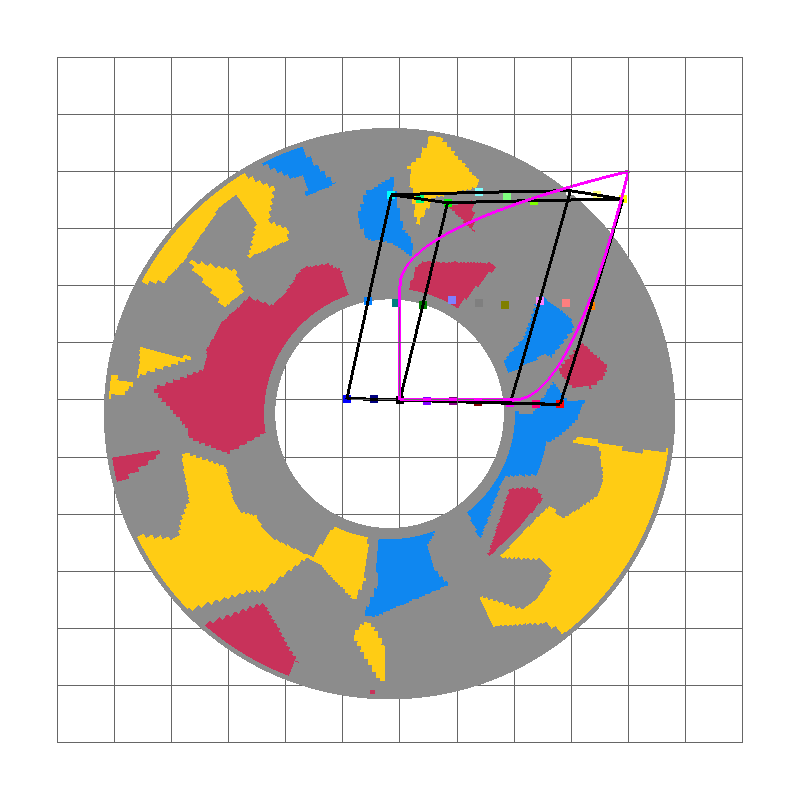}
      \\
      \textbf{a)}
      \includegraphics[keepaspectratio, width=0.3\textwidth,
      trim={3cm 4cm 3cm 1cm}, clip]
        {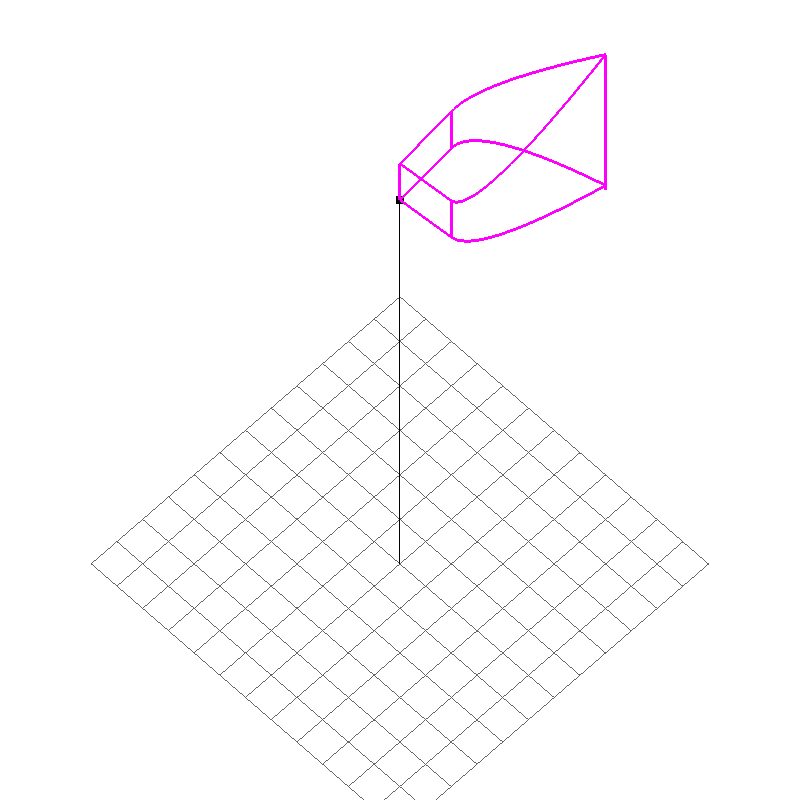}
      \textbf{b)}
      \includegraphics[keepaspectratio, width=0.3\textwidth,
      trim={3cm 4cm 3cm 1cm}, clip]
        {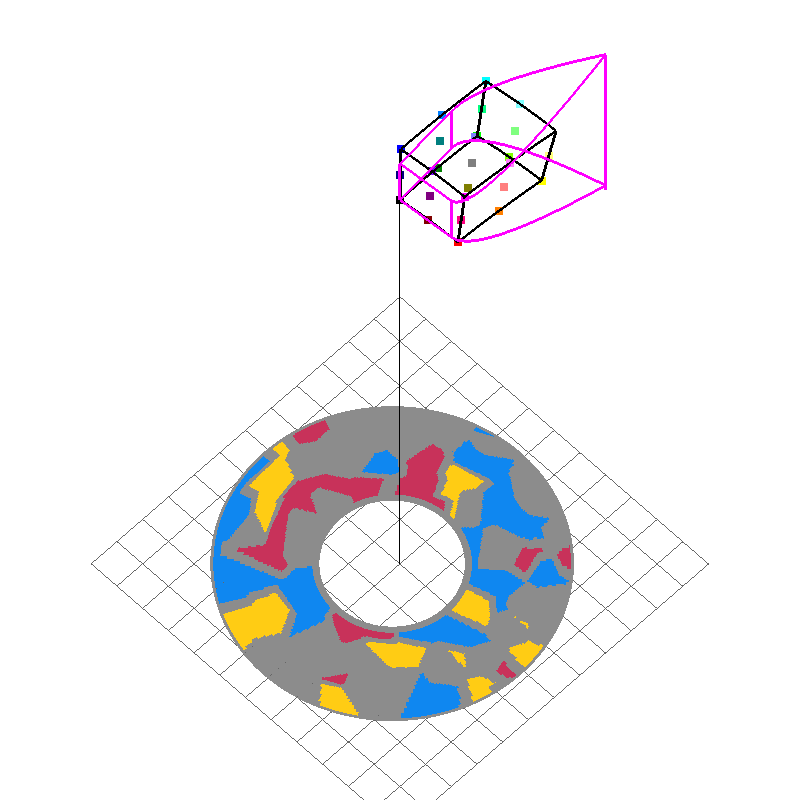}
      \textbf{c)}
      \includegraphics[keepaspectratio, width=0.3\textwidth,
      trim={3cm 4cm 3cm 1cm}, clip]
        {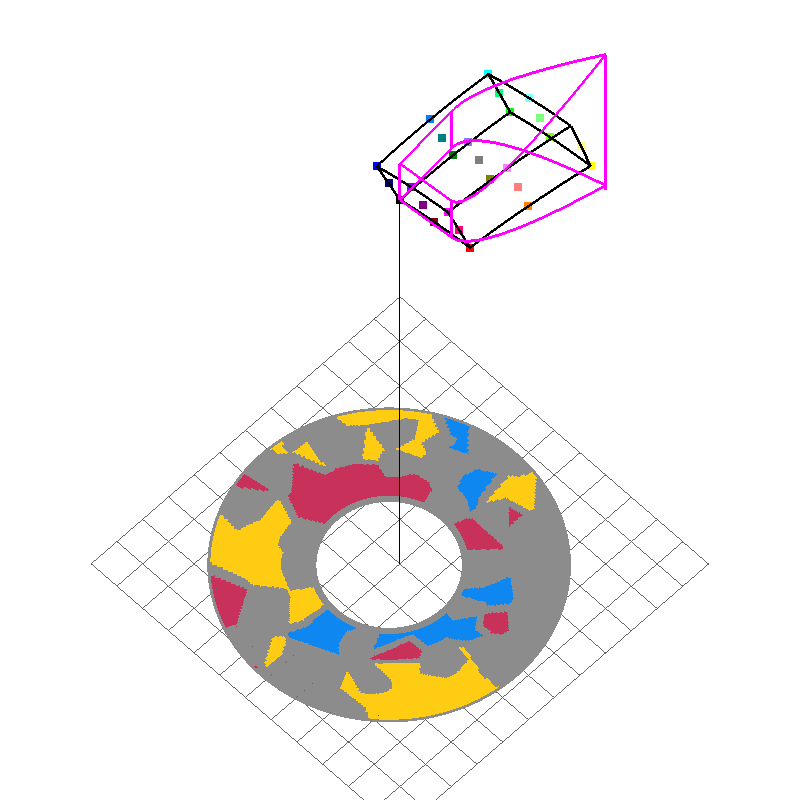}
        
    \caption{
      Third case study with a manually designed target workspace without being computed from a simulation using discretized cross-sections.
      \textbf{a)} The manually designed target workspace in top-down view and diagonal view,
      \textbf{b)} the best individual (cross-section and workspace) obtained through all iterations and methods and the target workspace (pink),
      \textbf{c)} the best individual (cross-section and workspace) obtained at the end of the worst performing iteration through all iterations and methods and the target workspace (pink)}
      \label{fig:optimized_design_c}
\end{figure}

We evaluate the performance of the genetic algorithm for the design of soft pneumatic actuators under investigation for three different target workspaces.
The first target workspace is derived from another soft pneumatic actuator with a random cross-section, generated using the Voronoi tessellation.
This target workspace and the cross-section used to generate it are shown in \autoref{fig:optimized_design_a}a).
The second target workspace is derived from a specific cross-section of a soft pneumatic actuator, designed by hand.
The manually designed cross-section, and the target workspace are shown in \autoref{fig:optimized_design_b}a).
The last target workspace has no specific cross-section underlying, and the target workspace is designed by hand.
The target workspace is shown in \autoref{fig:optimized_design_c}a).

\begin{figure}
  \centering

  \textbf{a)}
  \includegraphics[keepaspectratio, width=0.3\textwidth]
    {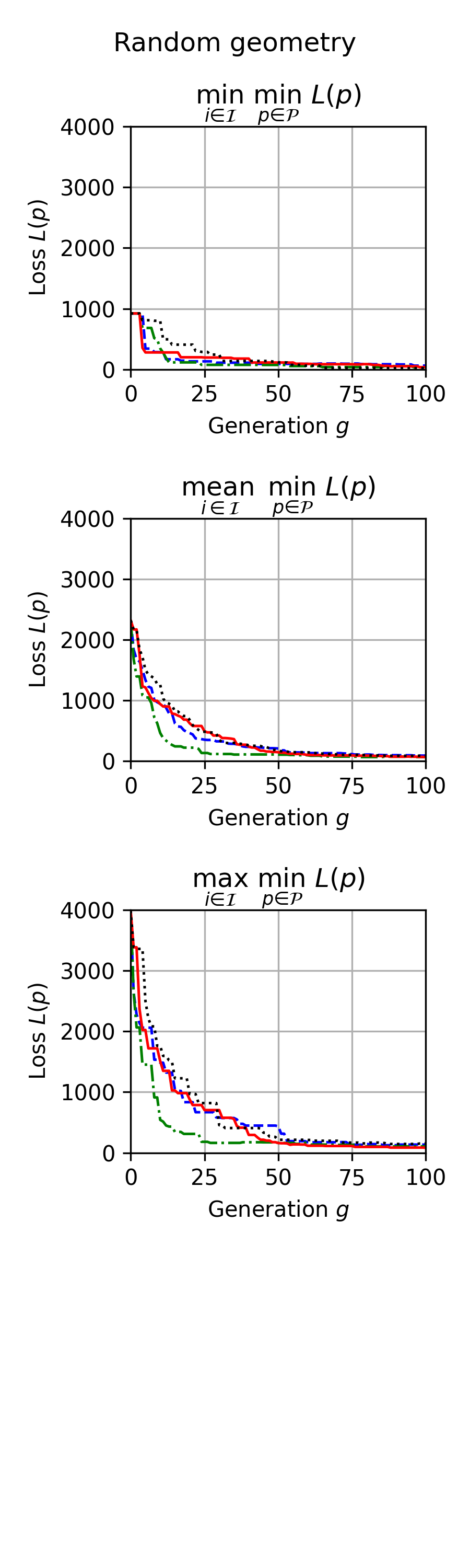}
  \textbf{b)}
  \includegraphics[keepaspectratio, width=0.3\textwidth]
    {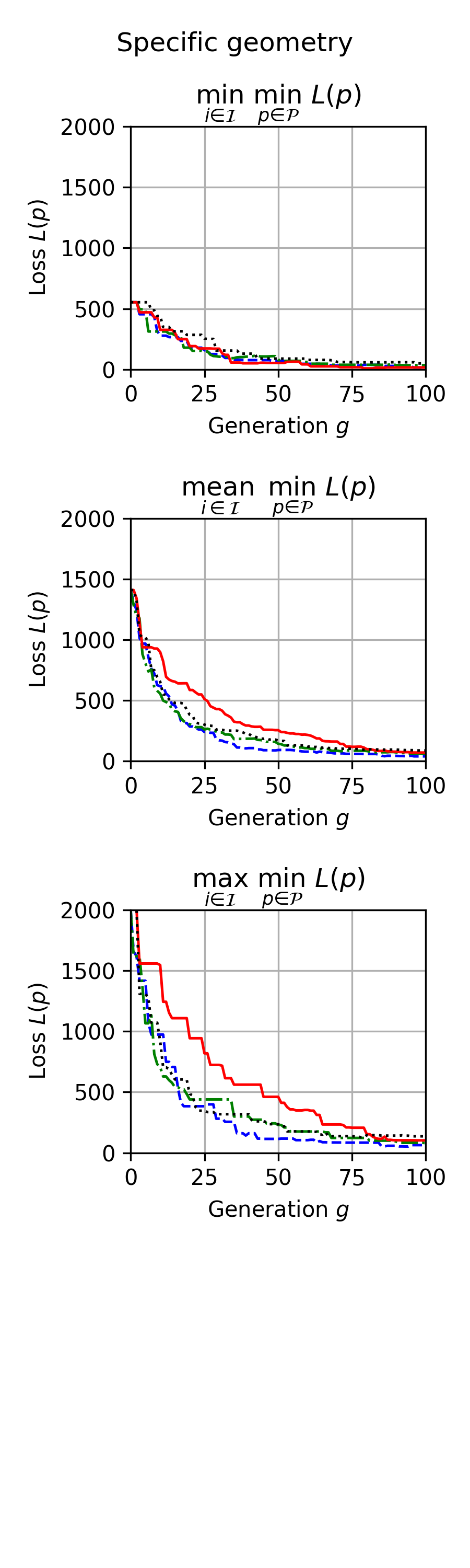}
  \textbf{c)}
  \includegraphics[keepaspectratio, width=0.3\textwidth]
    {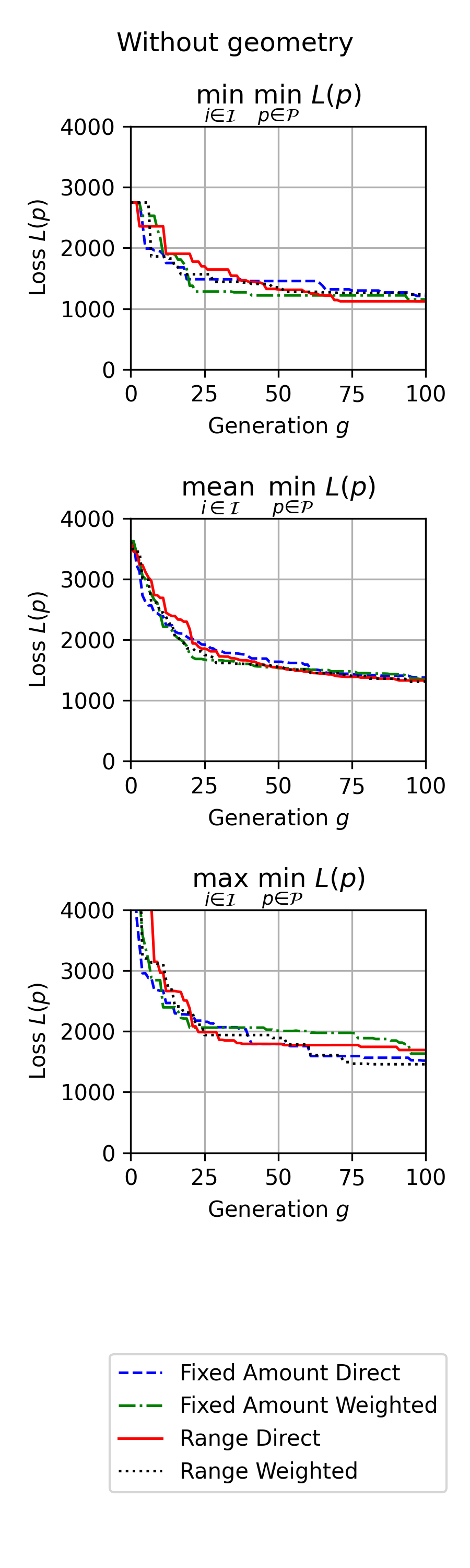}
    
  \caption{Minimum, mean, and maximum total loss of the best individual 
      in each generation over all iterations for each combination of
      recombination and mutation for the target workspace derived from
      \textbf{a)} a random cross-section, \textbf{b)} a specific cross-section, 
      and \textbf{c)} without a specific cross-section}
      \label{fig:loss}
\end{figure}

For each target workspace, four combinations of recombination and mutation methods are considered.
Each combination was run four times starting with different initial populations.
In the following, we refer to a single run as an iteration 
\begin{equation}
  i \in \{1, \dots, 4\} =: \mathcal{I}\ .
\end{equation}
The initial population between iterations is different, but reused for each combination of recombination and mutation and for each target workspace.
The considered combinations of recombination and mutation methods are
\begin{enumerate}
  \item[(a)] fixed recombination and direct mutation (FD),
  \item[(b)] fixed recombination and weighted mutation (FW),
  \item[(c)] range recombination and direct mutation (RD),
  \item[(d)] range recombination and weighted mutation (RW).
\end{enumerate}
The lower the loss $L$ of an individual the better this individual
is suited to reach the predefined target workspace.
Therefore, we used the lowest loss $\min_{p\in\mathcal{P}} L(p)$ of all possible individuals in this population as the evaluation metric for the performance of the genetic algorithm in each iteration and for each combination of recombination and mutation.
We plot the mean over all iterations, maximum over all iterations, and the minimum over all iterations of the loss of the best individual in each generation $\min_{p\in\mathcal{P}} L(p)$ for each combination of recombination and mutation in \autoref{fig:loss}.
Furthermore, in \autoref{tab:loss} we list the best loss at the end of the optimization, for all iterations, target workspaces and the four combinations of recombination and mutation methods.

\begin{table}[]
  \centering

  \caption{Summary of the loss $L$ of the best individual at 
  the end of the optimization for each combination of recombination and 
  mutation and for each target workspace}
  \label{tab:loss}

  \fontsize{8pt}{8pt}\selectfont
  \renewcommand{\arraystretch}{1.5}
  \begin{tabular}{|r||l|l|l|l||l|l|l|l||l|l|l|l|}
    \hline
   & \multicolumn{4}{c||}{Random geometry}
   & \multicolumn{4}{c||}{Specific geometry}
   & \multicolumn{4}{c|}{Without geometry}  \\
   It.
   & FD & FW & RD & RW 
   & FD & FW & RD & RW 
   & FD & FW & RD & RW \\ \hline \hline
   1 
   & 82.52 & 62.43 & 82.63 & 90.51
   & 15.07 & 34.22 & 93.06 & 97.55 
   & 1208.87 & 1321.27 & 1354.62 & 1236.93 \\
   2 
   & 81.80 & 32.39 & 35.81 & 90.54 
   & 51.82 & 56.04 & 15.21 & 48.81 
   & 1353.88 & 1633.49 & 1693.46 & 1458.76 \\
   3 
   & 60.98 & 105.04 & 144.40 & 19.49 
   & 63.10 & 79.06 & 102.04 & 66.83 
   & 1518.32 & 1298.97 & 1143.35 & 1268.10 \\
   4 
   & 134.04 & 62.85 & 67.42 & 152.52
   & 24.66 & 65.46 & 69.92 & 135.75 
   & 1416.24 & 1151.95 & 1119.73 & 1272.92 \\ \hline \hline
   Best 
   & 60.98 & 32.39 & 35.81 & \cellcolor{good} 19.49 
   & \cellcolor{good} 15.07 & 34.22 & 15.21 & 48.81 
   & 1208.87 & 1151.95 & \cellcolor{good} 1119.73 & 1236.93 \\ 
   Mean 
   & \cellcolor{bad} 89.84 & \cellcolor{good} 65.68 & 82.56 & 88.27
   & \cellcolor{good} 38.66 & 58.70 & 70.06 & \cellcolor{bad} 87.24
   & \cellcolor{bad} 1374.33 & 1351.42 & 1327.79 & \cellcolor{good} 1309.18 \\ 
   Worst 
   & 134.04 & 105.04 & 144.40 & \cellcolor{bad} 152.52
   & 63.10 & 79.06 & 102.04 & \cellcolor{bad} 135.75
   & 1518.32 & 1633.49 & \cellcolor{bad} 1693.46 & 1272.92 \\ \hline
  \end{tabular}
\end{table}

\newcommand{\figwidth}{0.24\textwidth}

For the target workspace generated from a random cross-section, this workspace and the cross-section are shown in \autoref{fig:optimized_design_a}a).
The best individual in this case is produced using the ranged-based recombination and weighted mutation method.
The best cross-section and the workspace of the best individual 
over all iterations and methods are displayed in \autoref{fig:optimized_design_a}b).
The cross-section and the corresponding workspace of the best individual
at the end of the iteration and method resulting in the worst result
are shown in \autoref{fig:optimized_design_a}c).

In the second case study, the target workspace is computed by defining a specific cross-section of the soft pneumatic actuator.
This cross-section and target workspace are shown in \autoref{fig:optimized_design_b}a).
The best result is obtained using the fixed amount recombination method and the direct mutation method.
The resulting cross-section and corresponding workspace are shown in \autoref{fig:optimized_design_b}b).
Furthermore, this combination of recombination and mutation methods performs the best on average across all iterations.
The worst result is generated using the range recombination and weighted mutation method in this case.
The cross-section and corresponding workspace for this individual are displayed in \autoref{fig:optimized_design_b}c) for comparison.

In the last case study, instead of starting from a cross-section of a soft pneumatic actuator and computing the target workspace, we directly specify the target workspace we want the soft pneumatic actuator to reach.
Since there are no restrictions to this target workspace it is clear that it would be easy to generate target workspace that would not be possible for a soft pneumatic actuator to reach.
We defined a target workspace that we believe is in the range of a plausible workspace, but still has some properties that would make it difficult for a soft pneumatic actuator to perfectly fit.
This target workspace is shown in \autoref{fig:optimized_design_c}a).
Interestingly, in contrast to the previous two case studies, the range-based recombination methods perform slightly better in this third case study.
The best individual is retrieved by the range-based recombination and direct mutation method in iteration four.
The cross-section and workspace of this individual is shown in \autoref{fig:optimized_design_c}b).
As already anticipated, given the possible unrealistic nature of the target workspace, this individual -- or, more precisely the computed workspace of the soft pneumatic actuator -- is not a perfect fit for the target workspace.
However, when interpreting the resulting workspace visually, position and alignment of the workspace seem to fit the given target workspace.
The ranged-based direct method not only yields the best individual across all iterations and combinations, but also yields the worst results for all iterations and combinations.
This individual is shown in \autoref{fig:optimized_design_c}c) with the cross-section and corresponding workspace of the soft pneumatic actuator.
In conclusion, in this case study all recombination and mutation methods perform more or less equally, with minor deviations.

\section{Discussion}
\label{sec:discussion}

As all four methods deliver similar results, the one with the shortest runtime should be selected.
Furthermore, one might argue that the range-based recombination methods could lead to cross-sections with undesirable properties due to having too many or too few feature points.

The recombination methods that use a fixed amount of feature points to exchange between the parents to form the offspring delivered better results than the range-based recombination methods over the four iterations when used to optimize a workspace derived from the cross-sectional geometry.
For the target workspace designed by hand, the range-based recombination method performed, on average, better.

For the mutation method, we do not see a benefit in computing the adjacent feature points for a selected feature point to determine the probability of the mutation outcome.
Therefore, we recommend the more naive approach of selecting the type of the non-selected feature point on the selected mutation edge.
This reduces the algorithm's runtime.

Since the design optimization method presented in this study could lead to ambiguity regarding the obtained cross-section for a given target workspace and the resulting deviations, a preprocessing step remains for the operator of this design assistant to select a manufacturable design.

\section{Conclusion \& Outlook}
\label{sec:conclusion_and_outlook}

In this study, we use black-box topology optimization to perform design optimization of slender soft pneumatic actuators.
We reduce the dimensionality of the problem to the design of a two-dimensional cross-section.
The performed optimizations identify possible cross-sectional designs for given target workspaces.
For target workspaces derived from the same cross-sectional structure, the target workspaces can be fitted quite well.
For an arbitrary target workspace, selected to be plausible, the optimized cross-sections are sufficient to produce a soft pneumatic actuator capable of approximating the target workspace.

We neglect some important aspects of the soft pneumatic actuator design to simplify both the design process and the required computational power.
We assume the soft pneumatic actuator to be torsion-stiff, which is justified, as long as the cross-section of the soft pneumatic actuator remains constant along its length and there are no external forces introducing torsional torques.
In future research, we intend to include the torsional stiffness of the soft pneumatic actuator and allow for the cross-section to vary along its length.

Another assumption made is the ideal stiffness of the actuator in radial direction.
Without a radial reinforcement, the actuator would balloon outward, leading to a change of pressurized area and, therefore a change in the pressure loads.
The ballooning must be considered in the formulation of the pressure forces or has to be mitigated by a radial reinforcement.

In future research, we plan to move the simulation to the real world, manufacture the optimized designs, and experimentally verify the workspace.
This verification should demonstrate how valid our assumptions are and how accurately a good design of a soft pneumatic actuator can be predicted using the genetic algorithm employed in this study.
For this manufacturing process, radial reinforcements need to be considered and integrate into the design and model.

As this study only focuses on soft pneumatic actuator without externally
applied loads, future research could consider the influence of external loads
and the arrangement of multiple actuators into a soft robotic system.
This could further increase the reachable workspace of the end-effector
and make target workspaces more feasible, like the workspace generated without being derived from an actual geometry.

By comparing the results obtained from the geometrically exact beam model to a finite element model, it could be assessed how well the beam model predicts the behavior of the soft pneumatic actuator.
As the beam model is less computational intensive than a finite element model, it is in our opinion a valid approach to tackle the design optimization
through genetic algorithms, as long as the soft pneumatic actuator has a slender structure.
However, they may suffer in accuracy compared to a full three-dimensional finite element model.

As genetic algorithms are gradient-free optimization methods and 
require a simulation of the soft pneumatic actuator for each
individual in the population, the computational cost of the genetic
algorithm is higher compared to gradient-based optimization methods.
In future research we intend to investigate the possibility of
differentiable physical simulators or generative artificial intelligence
to assist in the design optimization of soft pneumatic actuators
presented in this article.

The calculated workspaces $\mathcal{W}$ do not only contain the positions of the end-effector at different pressure levels, they further contain all possible trajectories in a quasi-static actuation.
Hence, instead of providing a target workspace, it might be sufficient to provide a set of target trajectories that must be included in the optimized workspace.
Additional constraints, such as tool forces or energy minimization, could then yield an optimal cross-sectional geometry for these trajectories.
Furthermore, as in the original mechanical context of topology optimization, the weight of the soft pneumatic actuator could also be considered as an objective, which should be reduced as much as possible.
We neglected the effects of dead weight from the actuator; however, it is clear that this load must be taken into account when applied to soft materials.

It should be noted that the designed soft robots are purely virtual and have neither been manufactured nor experimentally validated.
With this in mind, experimental validation is necessary to check if there are deviations between the simulated and experimental workspaces.

We believe that the workspace description used in this study and the focus on design optimizations given a specific target workspace, are steps in the direction of increasing the usability of soft pneumatic actuators.
The design assistant system reduces the expertise needed to describe the non-linear behavior and can shorten the overall design process time.

\section*{Acknowledgement}

We gratefully acknowledge the funding by the Deutsche Forschungsgemeinschaft 
(DFG, German Research Foundation) – 501861263 – SPP2353.
Furthermore, we would like to acknowledge Erik Faust for the suggestion
of using quadratic hexahedra as a method to describe the workspace of
the end-effector.

\nocite{*}
\printbibliography 

\end{document}